\documentclass[a4paper,fleqn,longmktitle]{cas-sc}
\usepackage{wrapfig}
\usepackage[numbers]{natbib}
\usepackage{setspace}
\usepackage{tikz}
\usepackage{graphicx}      
\usepackage{natbib}  
\usepackage{amsmath}
\usepackage{algorithm}
\usepackage{algorithmicx}
\usepackage{algpseudocode}
\usepackage[flushleft]{threeparttable}
\usepackage{lipsum}
\usepackage{multicol}
\usepackage{amssymb}
\usepackage{booktabs}
\usepackage{tikz}
\usepackage{ragged2e}
\usepackage{url}
\usepackage{tikz}
\usetikzlibrary{positioning, fit, arrows.meta, shapes}
\usepackage{textcomp}
\usepackage{subfigure}

\def\tsc#1{\csdef{#1}{\textsc{\lowercase{#1}}\xspace}}
\tsc{WGM}
\tsc{QE}
\tsc{EP}
\tsc{PMS}
\tsc{BEC}
\tsc{DE}
\newcommand*{\bigchi}{\mbox{\Large$\chi$}}

\begin{document}
\let\WriteBookmarks\relax
\def\floatpagepagefraction{1}
\def\textpagefraction{.001}
\shorttitle{}
\shortauthors{Agarwal et~al.}

\title [mode = title]{Explainability: Relevance based Dynamic Deep Learning Algorithm for Fault Detection and Diagnosis in Chemical Processes}                      
\tnotemark[1]


\author[1]{Piyush Agarwal}


\address[1]{Chemical Engineering Department, University of Waterloo, Ontario, Canada, N2L3G1}

\author[2]{Melih Tamer}
\address[2]{Manufacturing Technologies, Sanofi Pasteur, Toronto, Ontario, Canada, M2R3T4}

\author[3]{Hector Budman}
\cormark[1]

\address[3]{Chemical Engineering Department, University of Waterloo, Ontario, Canada, N2L3G1}



\doublespacing
\begin{abstract}
    The focus of this work is on Statistical Process Control (SPC) of a manufacturing process based on available measurements. Two important applications of SPC in industrial settings are fault detection and diagnosis (FDD). In this work a deep learning (DL) based methodology is proposed for FDD. We investigate the application of an explainability concept to enhance the FDD accuracy of a deep neural network model trained with a data set of relatively small number of samples. The explainability is quantified by a novel relevance measure of input variables that is calculated from a Layerwise Relevance Propagation (LRP) algorithm. It is shown that the relevances can be used to discard redundant input feature vectors/ variables iteratively thus resulting in reduced over-fitting of noisy data, increasing distinguishability between output classes and superior FDD test accuracy. The efficacy of the proposed method is demonstrated on the benchmark Tennessee Eastman Process.
\end{abstract} 




\maketitle

\section{Introduction}

The fourth industrial revolution also known as `Industry 4.0' and Big Data paradigm has enabled the manufacturing industries to boost its performance in terms of operation, profit and safety. The ability to store large amount of data have permitted the use of deep learning (DL) and optimization algorithms for the process industries. In order to meet high levels of product quality, efficiency and reliability, a process monitoring system is needed. The two important aspects of Statistical Process Monitoring (SPM) are fault detection and diagnosis (FDD). Normal operation of process plants can be detected by determining if the current state of the process is normal or abnormal where abnormal refers to a situation where a fault has occurred. This problem is referred to as ``Fault Detection". Following the detection of abnormality in the process, the next step is to diagnose the specific fault that has occurred. This step is referred to as ``Fault Diagnosis/ Classification". The presence of noise, correlation, non-linear process dynamics and high dimensionality of the process inputs greatly hinders the FDD mechanism in process plants. Previously, traditional multivariate statistical methods such as PCA, PLS and its variants \cite{wise1990theoretical,kaistha2001incipient,harmouche2014incipient,yuan2018multi} have been extensively used for fault detection and prognosis. However, the inherent non-linearity in the process pose challenges while using these linear methods and non-linear techniques can provide better accuracy. To this end, recent DL methods have shown considerable improvement over traditional methods. DL fault detection and classification techniques have been widely researched for applications in several engineering fields \cite{agarwal2019classification,agarwal2019deep}. In chemical engineering, machine learning techniques have been applied for the detection and classification of faults in the Syschem plant, which contains 19 different faults \cite{hoskins1991fault} and for the TEP problem. Beyond their application in the process industries Several studies on DL approaches have been conducted for the prevention of mechanical failures. For example DL models have been used for detecting and diagnosing faults present in rotating machinery \cite{janssens2016convolutional,jia2016deep}, motors \cite{sun2016sparse}, wind turbines \cite{zhao2018anomaly}, rolling element bearings \cite{gan2016construction,he2017deep} and gearboxes \cite{jing2017convolutional,chen2015gearbox}. Many DL studies have been recently conducted on TEP using DL models. \\

Although DL based models have better generalization capabilities, they are poor in interpretation abilities because of their black box nature. Using these methods it is difficult to identify the root cause of faults i.e. input variables that are most correlated to the occurrence of the faults by significantly deviating from their normal trajectories following the occurrence of the fault. Following the development of complex novel neural network architectures, there is an increasing interest in investigating the problems associated with DL models. For example, to understand how a particular decisions are made, which input variable/feature is greatly influencing the decision made by the DL-NN models, etc. This understanding is expected to shed light on why biased results can be obtained, why a wrong class is predicted with a higher probability in classification problems etc. Explainable Artificial Intelligence (XAI) is an emerging field of study which aims at explaining predictions of Deep Neural Networks (DNNs). Several different methods have been proposed in order to explain the predictions by assigning a relevance or contribution to each input variable for a given sample. Methodologies that are used to assign scores to each input feature with respect to a particular task can be classified into two class of methods: perturbation based methods and backpropagation based methods. Perturbation based methods perturb the individual input feature vectors (one by one) and estimate the impact on the output \cite{zeiler2014visualizing,zhou2015predicting} . On the other hand backpropagation methods are based on backward propagation of the probabilities calculated by Softmax output neurons in case of classification problem through different layers of the NN back to the input layer. Most perturbation methods are computationally expensive and often underestimate the relevance of the input features. To this end, different backward propagation methods have been proposed in the XAI literature for explaining the predictions such as Layer-wise Relevance Propagation (LRP) \cite{bach2015pixel}, LIME \cite{ribeiro2016should}, SHAP values \cite{lundberg2017unified}, DeepLIFT \cite{shrikumar2017learning}. LRP as a explainability technique has been successfully used in many different areas such as healthcare, audio source localization, biomedical domain and recently also in process systems engineering  \cite{montavon2019layer, agarwal2019classification, agarwal2019deep} and have been shown to perform better than both SHAP and LIME \cite{rios2020explaining}. In this work, we use LRP for explainability of the network by evaluating the relevance of input variables. LRP was proposed by Bach et al., 2015 \cite{bach2015pixel} to explain the predictions of DNNs by back-propagating the classification  scores from the output layer to the input layer.
In particular, for a specific output class c, the goal of LRP is to determine the relevance $R_{c}(x_i;f_{c})$ of the individual input variables/ feature vectors $\textbf{R}_{c}(\textbf{x};\textbf{f}_{c}) =[R_{c_1}(x_1;f_c), R_{c_2}(x_2;f_c), R_{c_3}(x_3;f_c),..,R_{c_i}(x_i;f_c),..,R_{c_{n}}(x_n;f_{c})] \in \mathbb{R}^{n}, i = 1,2,..,n$ of each input feature $x_i$ to the output $f_c(x)$.\\ 

\citeauthor{xie2015hierarchical}, \citeyear{xie2015hierarchical} proposed neural network based methodology as a solution for the diagnosis problem in the Tennessee Eastman simulation that combines the network model with a clustering approach. The classification results obtained by this method were satisfactory for most faults. Both Wang et al., \citeyear{wang2018generative} \cite{wang2018generative} and Spyridon et al. , \citeyear{spyridon2018generative} \cite{spyridon2018generative} proposed the use of Generative Adversarial Networks (GANs), as a fault detection scheme for the TEP. GANs are an unsupervised technique composed of a generator and a discriminator trained with the adversarial learning mechanism, where the generator replicates the normal process behavior and the discriminator decides if there is abnormal behavior present in the data. This unsupervised technique can detect changes in the normal behavior achieving good detection rates. Lv et al., \citeyear{lv2016fault} \cite{lv2016fault} proposed a stacked sparse autoencoder (SSAE) structure with a deep neural network to extract important features from the input to improve the diagnosis problem in the Tennessee Eastman simulation. The diagnosis results applying this DL technique showed improvements compared to other linear and non-linear methods. To account for dynamic correlations in the data, Long Short Term Memory (LSTM) units have been recently applied to the TEP for the diagnosis of faults \cite{zhao2018sequential}. A model with LSTM units was used to learn the dynamical behaviour from sequences and batch normalization was applied to enhance convergence. An alternative to capture dynamic correlations in the data is to apply a Deep Convolutional Neural Networks (DCNN) composed of convolutional layers and pooling layers \cite{wu2018deep}.\\ 

The fault detection problem in the current work is formulated as a binary classification problem where the objective is to classify whether the current state of the process plant is normal or abnormal while the fault diagnosis problem is formulated as a multi-class classification problem to identify the type of fault. Then, the application of the concept of explainability of Deep Neural Networks (DNNs) is explored with its particular application in FDD problem. While the explainability concept has been studied for general DNN based models as mentioned above, it has not been investigated before in the context of FDD problems. In this work the relevance of input variables for FDD are interpreted using LRP and the irrelevant input variables’ for the supervised classification problem are discarded. It is shown that the resulting pruning of the input variables results in enhanced fault detection as well as fault diagnosis test accuracy. Lastly, we show that the use of a Dynamic Deep Supervised Autoencoder (DDSAE) NNs along with the pruning of the network for both fault detection and diagnosis further improves the overall classification ability as compared to other methods reported before.\\

To conduct a fair comparison of the proposed algorithm to previously reported methods, careful attention should be given to the data used as the basis for comparison.  For example, there is a vast literature on FDD for the TEP problem that uses differing amounts of data. In this work, we have used a standard dataset as a basis for comparison which further challenges the training of DL models and accuracy of FDD with a DL model and that has been used for comparison in other studies. The proposed DL based detection method with Deep Supervised Autoencoder (DSAE) or Deep Dynamic Supervised Autoencoder (DDSAE) is compared to several techniques: linear Principal Component Analysis (PCA) \cite{zhang2009enhanced,yin2012comparison,lau2013fault,shams2010fault}, Dynamic Principal Component Analysis (DPCA) \cite{chiang2000fault,yin2012comparison,ku1995disturbance,rato2013fault,odiowei2009nonlinear} , Independent Component Analysis (ICA) \cite{hsu2010novel} and with two other recently reported methods that use DL models based on Sparse Stacked Autoencoder NNs (SAE-NN) \cite{lv2016fault} and Convolutional NN (CNN)) CNN\cite{chadha2017comparison} for the same data set. For the Fault Diagnosis problem, the proposed method is compared with Support Vector Machines (SVM) \cite{kulkarni2005knowledge,chiang2004fault,mahadevan2009fault}, Random Forest, Structure SVM, and sm-NLPCA (architecture used: Stacked Autoencoder). It will be shown that the proposed relevance based method with DSAE or DDSAE networks significantly increases the average fault detection and diagnosis accuracy over other methods.\\

The paper is organized as follows. The mathematical modelling tools including basic Autoencoder (AE), Deep Supervised Autoencoder (DSAE), its dynamic version Dynamic Deep Supervised Autoencoder (DDSAE) neural networks and Layerwise Relevance Propagation (LRP) for computing the relative importance of input variables for explaining the predictions of DNNs are introduced in Section 2. The developed methodology for both Fault Detection and Diagnosis (FDD) is presented in Section 3. The application of the proposed method to the case study of Tennessee Eastman Process and comparisons to other methods are presented in Section 4 followed by conclusions presented in Section 5.

\section{Preliminaries}
This section briefly reviews the fundamentals of an Autoencoder (AE-NNs) and a Supervised Autoencoder Neural Networks (SAE-NNs) models. 

\def\layersep{2.5cm}
{\tiny\begin{figure}[]
\centering
\begin{tikzpicture} [shorten >=1pt,->,draw=black!50, node distance=\layersep]
    \tikzstyle{every pin edge}=[<-,shorten <=1pt]
    \tikzstyle{neuron}=[circle,fill=black!25,minimum size=17pt,inner sep=0pt]
    \tikzstyle{input neuron}=[neuron, fill=green!50];
    \tikzstyle{output neuron}=[neuron, fill=red!50];
    \tikzstyle{hidden neuron}=[neuron, fill=blue!50];
    \tikzstyle{annot} = [text width=4em, text centered]

    \foreach \name / \y in {1,...,5}
        \node[input neuron, pin=left:${x}_\y$] (I-\name) at (0,-\y) {};

    \foreach \name / \y in {1,...,3}
        \path[yshift=-0.75cm]
            node[hidden neuron] (H-\name) at (\layersep,-\y cm) {};

    \foreach \name / \y in {1,...,5}
    \node[output neuron, pin = {[pin edge={->}]right:$\hat{x}_\y$}] (O-\name) at (5,-\y cm) {};
    \foreach \source in {1,...,5}
        \foreach \dest in {1,...,3}
            \path (I-\source) edge (H-\dest);

    \foreach \source in {1,...,3}
        \foreach \dest in {1,...,5}
            \path (H-\source) edge (O-\dest);

    \node[annot,above of=H-1, node distance=1.8cm] {Latent Space\\ (\textbf{z})};
    \node[annot,above of=I-1,node distance=1.8cm] {Input layer\\ (\textbf{x})};
    \node[annot,above of=O-1, node distance = 1.2cm]{Output layer\\ ($\hat{\textbf{x}}$)};
    \node[annot,right of=I-3,node distance=1.5cm] {\footnotesize{}${f_e(\textbf{x})}$};
    \node[annot,left of=O-3,node distance=1.4cm] {\footnotesize{}${f_d(z)}$};
\end{tikzpicture}
\caption{Traditional single layer Autoencoder Neural Network (AE-NN)}
\label{autoencoder}
\end{figure}
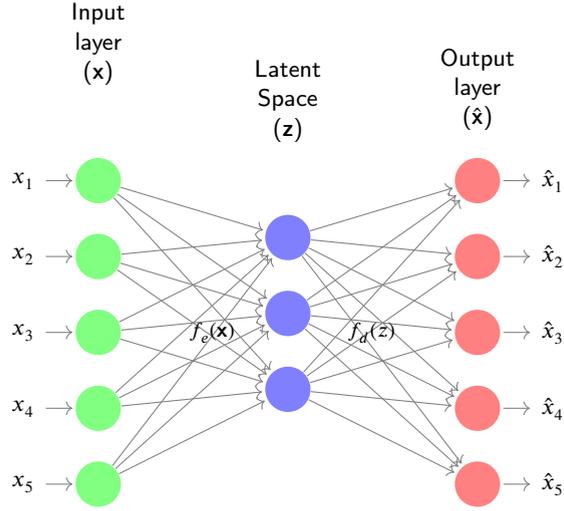}

\subsection{Autoencoder Neural Networks (AE-NNs)}
A traditional AE-NN is a neural network model composed of two parts: encoder and decoder, as shown in Figure \ref{autoencoder}. An AE is trained in an unsupervised fashion to extract underlying patterns in the data and to facilitate dimensionality reduction. The encoder is trained so as to compress the input data onto a reduced latent space defined within a hidden layer and the decoder uncompresses back the hidden layer outputs into the reconstructed inputs. 
Let us consider the inputs to an AE-NN $\mathbf{X} = [\mathbf{x}_1 ~\mathbf{x}_2~ \mathbf{x}_3 \hdots \mathbf{x}_{N}]^T \in \mathbb{R}^{N \times d_x}$, then the operation performed by the encoder for a single hidden layer between the input variables to the latent space $\mathbf{z} \in \mathbb{R}^{d_z}$ variables (latent variables) for time sample $i$ can be represented as follows:
\begin{align}
    \textbf{z}_i = f_e(\textbf{W}_{e}\textbf{x}_i + \textbf{b}_e)
\end{align}
where $f_e$ is a chosen non-linear activation function for the encoder, $\textbf{W}_e \in \mathbb{R}^{d_z \times d_x}$ is an encoder weight matrix, $\textbf{b}_e \in \mathbb{R}^{d_z}$ is a bias vector. The decoder reconstructs back the input variables from the feature or latent space $\textbf{z}_i \in \mathbb{R}^{d_z}$ as per the following operation follows:
\begin{align}
    \hat{\textbf{x}}_i = f_d(\textbf{W}_{d}\textbf{z}_i + \textbf{b}_d)
\end{align}
where $f_d$ is a chosen activation function for the decoder, $\textbf{W}_d \in \mathbb{R}^{d_x \times d_z}$ and $\textbf{b}_d \in \mathbb{R}^{d_x}$ is a decoder weight matrix and a bias vector respectively. The `$\tanh$' function is used for both transforming the inputs into the latent variables and for reconstructing back the inputs from the latent variables as an example here. The AE-NN is trained based on the following minimization problem:
\begin{align}
    l_{AE}(\textbf{x},\textbf{W}_d\textbf{W}_e\textbf{x}) = \frac{1}{2N}||\textbf{x} - \hat{\textbf{x}}||_2^2 = \frac{1}{2N}\sum_{s = 1}^{N}\big(\textbf{x}_s - \hat{\textbf{x}}_s\big)^2
    \label{AE_lossfunction}
\end{align}
where $N$ is the number of samples. 

\begin{figure}[]
    \centering

    \includegraphics[scale = 0.35 ]{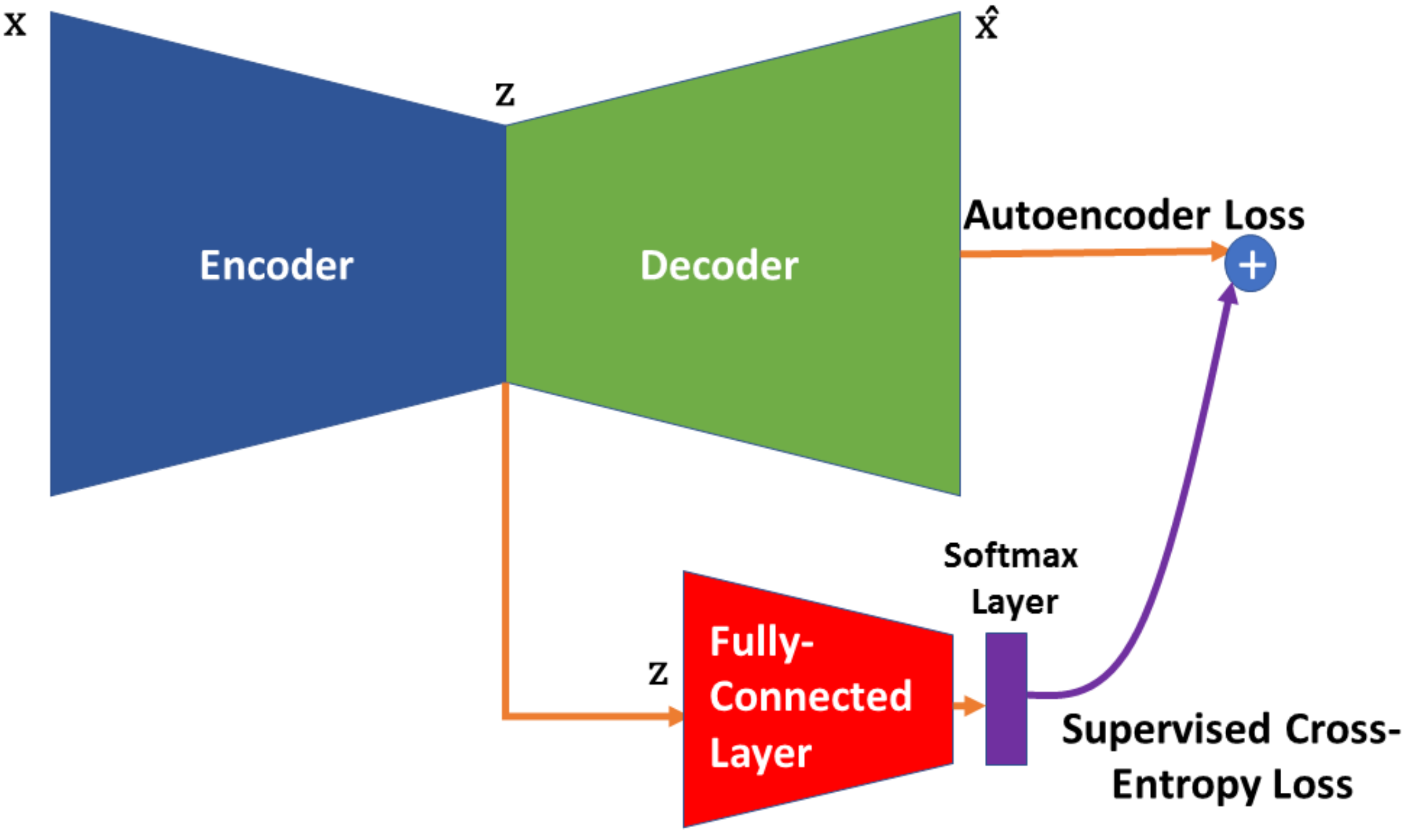}
    \caption{Schematic of a single layer Supervised Autoencoder Neural Network (SAE-NN)}
    \label{Schematic SAE}
\end{figure}
\subsection{Deep Supervised Autoencoder Classification Neural Networks (DSAE-NNs)}
 The overall goal is to learn a function that predicts the class labels in one-hot encoded form $\textbf{y}_{i} \in \mathbb{R}^m$ from inputs $\textbf{x}_i \in \mathbb{R}^{d_x}$. The operation performed by the encoder for a single hidden layer between the input variables to the latent variables $\mathbf{z}_i \in \mathbb{R}^{d_z}$ can be mathematically described as follows:
\begin{align}
    \textbf{z}_i = f_e(\textbf{W}_{e}\textbf{x}_i + \textbf{b}_e)
\end{align}
The latent variables are used both to predict the class labels and to reconstruct back the inputs $\textbf{x}$ as follows:
\begin{align}
    \hat{\textbf{x}_i} = f_d(\textbf{W}_{d}\textbf{z}_i + \textbf{b}_d) \\
    \hat{\textbf{y}_i} = f_c(\textbf{W}_c\textbf{z}_i + \textbf{b}_c)
\end{align}
where $f_c$ is a non-linear activation function applied for the output layer. $\textbf{W}_c \in \mathbb{R}^{m \times d_z}$ and $\textbf{b}_c \in \mathbb{R}^{m}$ are output weight matrix and bias vector respectively.
The training of an Deep Supervised Autoencoder Neural Network (DSAE-NN) model, schematically shown in Figure \ref{Schematic SAE}, is based on the minimization of a weighted sum of the reconstruction loss function and the supervised classification loss corresponding to the first and second terms in  (Equation (\ref{SAE_lossfunction})) respectively. The reconstruction loss function in Equation (\ref{SAE_lossfunction}) is ensuring that the estimated latent variables are able to capture the variance in the input data while the classification loss is ensuring that only those non-linear latent variables are extracted that are correlated with output classes. Mean squared error function is used as a reconstruction loss and softmax cross-entropy as the classification loss.
\begin{align} \nonumber l_{DSAE} &= \lambda_1 \sum_{s = 1}^{N}L_{r}^{s}(\textbf{x}_s,\textbf{W}_d\textbf{W}_e\textbf{x}_s) +  \sum_{s = 1}^{N}L_{p}^{s}(\textbf{W}_c \textbf{W}_e\textbf{x}_s,\textbf{y}_s)\\ \nonumber
 &=  \frac{\lambda_1}{N}|| \textbf{x}_s - \hat{\textbf{x}}_s||_2^2 + \frac{1}{N}\sum_{s=1}^{N}\sum_{c=1}^{m} -y_{s,c}log(p_{s,c})\\ 
 &=  \frac{1}{N}\bigg[\lambda_1||\textbf{x}_s - \hat{\textbf{x}}_s||_2^2 +\sum_{s=1}^{N}\sum_{c=1}^{m} -y_{s,c}log(p_{s,c})\bigg]    
\label{SAE_lossfunction}
\end{align}
\begin{align}
    p_{s,c} &= \frac{e^{(\hat{y_{s,c}})}}{\sum_{c = 1}^{m}e^{(\hat{y_{s,c}})}}
\end{align}
where $\lambda_1$ is the weight for the reconstruction loss $L_r$, $m$ is the number of classes, $y_{s,c}$ is a binary indicator (0 or 1) equal to 1 if the class label $c$ is the correct one for observation $s$ and 0 otherwise, $\hat{y_{s,c}}$ is the non-normalized log probabilities and $p_{s,c}$ is the predicted probability for a sample $s$ of class $c$. Moreover, to avoid over-fitting, a regularization term is added to the objective function in Equation \ref{SAE_lossfunction}. Hence, the objective function for Deep Supervised Autoencoder NNs used for Fault Detection (number of classes $m = 2$, normal or faulty) is as follows:
\begin{align}
    \min_{\textbf{W}} l_{DSAE} = \min ~~\frac{1}{N}\bigg[\lambda_1||\textbf{x}_s - \hat{\textbf{x}}_s||_2^2 +\lambda_2\sum_{s=1}^{N}\sum_{c=1}^{m} -y_{s,c}log(p_{s,c}) + \lambda_3\sum_{L}\sum_{k}\sum_{j}\textbf{W}_{{kj}}^{2 [L]} \bigg] 
    \label{DSAE_objective_function}
\end{align}

where $\textbf{W}_{kj}^{[L]}$ are the weight matrices for each layer $L$ in the network (L = 1 in this example) and the weights on the individual objective functions  $\lambda_1, \lambda_2, \lambda_3$ are chosen using validation data.

\subsection{Dynamic Deep Supervised Autoencoder Classification Neural Networks (DDSAE-NNs)}
The static DSAE-NN presented above assumes that the sampled data are independent to each other, and hence, temporal correlations are ignored. To account for the correlations in time between the data samples, a Dynamic Deep Supervised Autoencoders (DDSAE) model has been proposed by using a dynamic extension matrix. Accordingly, the original DSAE-NN model can be extended to take into account auto-correlations in time correlated data by augmenting each sample vector with the previous $l$ observations and stacking the data matrix with the resulting vectors, each corresponding to different time intervals. \\

The dynamic augmentation of the input data matrix  $\textbf{X}_{D}$ by stacking previous $l$ observations to input data matrix $\textbf{X}$ is as follows:

\begin{equation}
    \textbf{X}_{D} = \big[ \textbf{x}_{l+1}^{D}~ \textbf{x}_{l+2}^{D}~ \textbf{x}_{l+3}^{D}\hdots \textbf{x}_{N}^{D}\big]^{T} \in \mathbb{R}^{(N-l) \times ((l+1)d_x)}
    \label{augmented_matrix}
\end{equation}
where $\textbf{x}_{n}^{D} = [\textbf{x}_{n}~ \textbf{x}_{n-1}~ \textbf{x}_{n-2} \hdots \textbf{x}_{1}]$, where $\textbf{x}_{n}$ is a $\mathbb{R}^{d_x}$ dimensional vector of all the input feature vectors/ variables. The different time window length is chosen to build DDSAE models, in which the best classification performance is the final time window length. The following objective function ( Equation \ref{DDSAE_objective_function})  is minimized with training data $\textbf{X}_{D}^{N-l+1} = \{\textbf{x}\}_{i = 1}^{N-l+1}, \textbf{y}_{D} = \{y\}_{i = 1}^{N-l+1}$, where $N-l+1$ is the total number of samples:

\begin{align}
    \min_{\textbf{W}} l_{DDSAE} = \min ~~\frac{1}{(N-l+1)}\bigg[\lambda_1||\textbf{x}^{D}_s - \hat{\textbf{x}}^{D}_s||_2^2 +\lambda_2\sum_{s=1}^{N-l+1}\sum_{c=1}^{m} -y_{s,c}log(p_{s,c}) + \lambda_3\sum_{L}\sum_{k}\sum_{j}\textbf{W}^{2[L]}_{{kj}} \bigg] 
    \label{DDSAE_objective_function}
\end{align}

Note that the number of samples for the augmented dynamic matrix for the training data decreases as compared to the static DSAE case.

\subsection{{Layer-wise Relevance Propagation (LRP)}}

Layer-wise Relevance Propagation (LRP) was introduced by \citeauthor{bach2015pixel}, (\citeyear{bach2015pixel})\cite{bach2015pixel} to assess the relevance of each input variable or features with respect to outputs using a trained NN. It is based on a layer-wise relevance conservation principle where each relevance $[R_{c}(x_i;f_{c})]_j,~~ i = 1,2,...,n$, where $n$ is the number of input variables/ feature vectors ($x_i,~~ i = 1,2,...,n$) for a $j^{th}$ sample where $j = 1,2,...,N$, where $N$ is the total number of samples in the training dataset, is calculated by propagating the output scores for a particular task $c$ towards the input layer of the network. Previously, it has been used in the area of health-care for attributing a relevance value to each pixel of an image to explain the relevance in a image classification task using DNNs \cite{yang2018explaining}, to explain the predictions of a NN in the area of sentiment analysis \cite{arras2017explaining}, to identify the audio source in reverberant environments when multiple sources are active \cite{perotin2019crnn}, and to identify EEG patterns that explain decisions in brain-computer interfaces \cite{sturm2016interpretable}. In process systems engineering LRP has been recently applied by the authors for the first time for FDD problems. The method was used for identifying relevant input variables and pruning irrelevant input variables (input nodes) with respect to a specific classification task for both Multilayer Perceptron (MLP) NN and Long-Short Term Memory (LSTM) NN by \citeauthor{agarwal2019classification}, (\citeyear{agarwal2019classification}) \cite{agarwal2019classification,agarwal2019deep}.\\

To compute the relevance of each input variable $x_i, i = 1,2,...,n$ for the DSAE-NNs and DDSAE-NNs models (used in this work), are trained for both fault detection and fault classification using the training dataset $\bigchi:\{\textbf{X}^{l},\textbf{y}^{l}\}$ and $\bigchi_{D}:\{\textbf{X}_{D}^{l},\textbf{y}_D^{l}\}$ ($\bigchi_D$: dynamic version of $\bigchi$ ) respectively. Subsequently, the score value $f_c$ of the corresponding class for the $j^{th}$ sample is back-propagated through the network towards the input. Depending on the nature of the connection between layers and neurons, a layer-by-layer relevance score is computed for each intermediate lower-layer neuron. Different LRP rules have been proposed for attributing relevance for the input variables. In this work, we use the $\epsilon$ epsilon rule\cite{bach2015pixel} for computing relevances that are given as follows:
\begin{align}
    R_{l\leftarrow u} = \sum_{u}\frac{x_{l}w_{lu}}{\sum_{j}x_{l}w_{lu} +\epsilon}R_u
\end{align}
where $\epsilon$ is used to prevent numerical instability when $z_u$ is close to zero and $w_{ul}$ are the weights connecting lower layer neurons $l$ and upper-layer neurons $u$. As $\epsilon$ becomes larger, only the most salient explanation factors survive the absorption. This typically leads to explanations that are sparser in terms of input features and less noisy \cite{montavon2019layer}. Relevances $\textbf{R}_{c}(\textbf{x};\textbf{f}_{c})_j =[R_{c_1}(x_1;f_c), R_{c_2}(x_2;f_c),...,R_{c_{n}}(x_n;f_{c})]_j \in \mathbb{R}^{n}, ~i = 1,2,..,n$ for each input feature $x_i$ are calculated for the $j^{th}$ sample with respect to a classification task $c$ in the training dataset $\bigchi$ or $\bigchi_D$. Since the goal is to prune the irrelevant input features DNNs based on estimated relevance scores using LRP, it is important to average the relevance scores over all the correctly classified samples in the training dataset. Therefore, the final input relevances with respect to the overall classification task $c$ can be calculated as follows:

\begin{align}
\textbf{R}_c = \frac{1}{N_c} \sum_{j = 1}^{N_c} \textbf{R}_c(\textbf{x};\textbf{f}_c)_j
\label{average_relevance_score}
\end{align}

where $N_c$ is the number of correctly classified samples in the training dataset. Furthermore, the least relevant input features are pruned based on the average relevance scores for all input variables calculated with Equation \ref{average_relevance_score}. In practise a threshold of $\lambda \times \max(\textbf{R}_c)$ is chosen to prune the irrelevant variables where $\lambda$ is an hyper-parameter that is determined by using the validation dataset (heuristically $\lambda$ is chosen as 0.01 as the starting value). Relevance of input variables below the threshold are removed from the dataset and the network is re-trained until the same or higher validation accuracy is achieved. It is to be noted that the DNN has to be re-trained with the set of remaining input variables after pruning and the testing accuracy increases with successive iterations as shown later in Section 4. For the dynamic augmented input matrix $\bigchi_D$ shown in Equation 10, the final input relevances (combining effects of individual lagged variables) with respect to the overall classification task $c$ is:\\

\begin{align}
    \textbf{R}_c = \frac{1}{N_c} \sum_{j = 1}^{N_c}\sum_{i = 1}^{l+1} |R_{c_{i}}(x_i;f_c)|_j
\end{align}
where $l$ is the number of time lag window included in the dataset $\textbf{X}^l_D$.

\section{Fault Detection and Diagnosis Methodology based on DSAE-NNs and DDSAE-NNs}

Both the Deep Supervised Autoencoder NN (DSAE-NN)and Dynamic Deep Supervised Autoencoder NN (DDSAE-NN) are used for FDD and are the basis for the explainable-pruning based methodology presented in the previous section. The proposed fault detection algorithm is first used to extract deep features to detect if the process is operating in a normal or faulty region. Then, a fault diagnosis algorithm is applied in case the sample indicates faulty operation to identify the particular fault and possible root-cause of the occurring fault in the process using an DDSAE-NN. Since the latter is iteratively trained by using the LRP based pruning procedure that provides explainability of input variables the resulting DDSAE-NN model will be referred to as xDDSAE-NN. 

\subsection{Fault Detection Methodology}
First, a DSAE-NN is trained using the training data ($\textbf{X}^{l}, \textbf{y}^{l}$). The fault detection process is formulated as a binary classification problem. Often, this binary classification task for Fault Detection is susceptible to a `class imbalance problem' because of the unequal distribution of classes in the training dataset. For example, the number of training samples for the normal operating region may be far less than the samples for abnormal operating region or vice-versa. To address this class imbalance problem an extra weight $\delta$ is introduced in the loss functions in Equation \ref{DSAE_objective_function1} and Equation \ref{DDSAE_objective_function1} is as follows:
\begin{align}
    \min_{\textbf{W}} l_{DSAE} = \min ~~\frac{1}{N}\bigg[\lambda_1\sum_{i = 1}^{N}||\textbf{x}_s - \hat{\textbf{x}}_s||_2^2 -\lambda_2\sum_{s=1}^{N} \big(\delta y_{s,1}\log(p_{s,1}) + y_{s,2}\log(p_{s,2}) \big)+ \lambda_3\sum_{L}\sum_{k}\sum_{j}\textbf{W}_{kj}^{2[L]} \bigg] 
    \label{DSAE_objective_function1}
\end{align}

\begin{align}
    \min_{\textbf{W}} l_{DDSAE} = \min ~~\frac{1}{N}\bigg[\lambda_1\sum_{i = 1}^{N}||\textbf{x}^{D}_s - \hat{\textbf{x}}^{D}_s||_2^2 -\lambda_2\sum_{s=1}^{N} \big(\delta y_{s,1}\log(p_{s,1}) + y_{s,2}\log(p_{s,2}) \big) + \lambda_3\sum_{L}\sum_{k}\sum_{j}\textbf{W}_{{kj}}^{2[L]} \bigg] 
    \label{DDSAE_objective_function1}
\end{align}

For example, if there are more data samples of faulty operation than samples for normal operation higher weights would be assigned to the samples belonging to the normal operating region class. The value of $\delta$ dictates a trade-off between false positives and true negatives and is considered as an additional hyper-parameter to the model that is ultimately chosen using the validation data-set. Initially the DSAE-NN is trained on the training dataset ${\textbf{X}^{l},\textbf{y}^{l}}$ using all the input-variables. The best performing model is chosen using a validation dataset $\textbf{X}^{v},\textbf{y}^{v}$. Then, the LRP is implemented to explain the predictions of the chosen DSAE-NN with a set of hyper-parameters by computing the relevance of each input variable. The irrelevant input feature vectors are discarded based on a threshold as described in Section 2.4. Subsequently, the eXplainable DSAE (xDSAE) neural network is re-trained using the reduced training dataset $\{\textbf{X}^{l},\textbf{y}^{l}\} \rightarrow \{\textbf{X}^{l}_{r},\textbf{y}^{l}_{r}\}$ at each iteration. The premise for reducing the dimensionality of the input data by discarding less relevant inputs is that the information content can often be represented by a lower dimensional space, implying that only a few fundamental variables are sufficient to account for the variation in the data that are most informative about the identification of faults and normal regions. Once all the relevant input variables that are significant to the classification task are chosen, an eXplainable DDSAE-NN (xDDSAE-NN) is trained with the remaining inputs and the reduced training data matrix $\textbf{X}^{l}_{r}$ is augmented with the lagged variables of the remaining input variables $\textbf{X}^{l~D}_{r}$. The process of discarding input feature vectors is iterative and xDDSAE-NN is iteratively retrained using the validation dataset. This approach has multiple advantages over other reported methods used for fault detection as follows: 

\begin{enumerate}
    \item Improvement in test classification accuracy.
    \item Identification of an eXplainable empirical model
    \item Synthesis of a smaller network with fewer parameters
\end{enumerate}

\subsection{Fault Diagnosis Methodology}
    After detecting that the process has deviated from the normal operation and a fault has occurred, it is desired to diagnose the type of fault. For fault diagnosis, a similar methodology to the one used for fault detection is applied for the classification of the type of fault. First, the static DSAE is used to extract deep features and predict the type of fault in a process plant. For this task one-hot encoded outputs are utilized as the labels for training the model. Initially, the DSAE-NN is trained on the training dataset ${\textbf{X}^{l},\textbf{y}^{l}}$ using all the input-variables. The best performing model is chosen using a validation $\textbf{X}^{v},\textbf{y}^{v}$. LRP is subsequently implemented to explain the predictions of the selected DSAE-NN by computing the relevance of each input variable. The irrelevant features are removed by comparing the relevances to a threshold. Then an xDSAE NN is trained using the reduced training dataset {$\textbf{X}^{l}_{r},\textbf{y}^{l}_{r}$} by successive iterations of pruning of irrelevant inputs and model re-training until the relevance of all the remaining input variables are above the threshold. Since data collected from chemical processes have strong dynamic/temporal correlations, the input data matrix $\textbf{X}$ is augmented with observations at $l$ previous time steps for each input feature dimension (refer Equation \ref{augmented_matrix}) and a DDSAE-NN is trained. The iterative procedure of discarding input variables from the reduced dynamic matrix $\textbf{X}^{l~D}_{r}$ is implemented and pruning and re-training is applied as long as validation accuracy continue to increase after discarding features. The decision of adding lagged variables only to the remaining input variables of the final iteration of xDSAE model is justified by the fact that the input variables that were eliminated do not have an instantaneous effect of $x_k$ on fault detection. Then, since the pruned input variables at current time are auto-correlated in time to previous values ($x_k \propto f(\textbf{x}_{k-1},\textbf{x}_{k-2},..,\textbf{x}_{k-n})$), if current values are not correlated to the model outputs then their corresponding previous values (lagged variables) are also not correlated to these outputs. 
    
\subsection{Proposed Methodology for FDD}
\begin{figure}
    \centering
    \includegraphics[scale = 0.8]{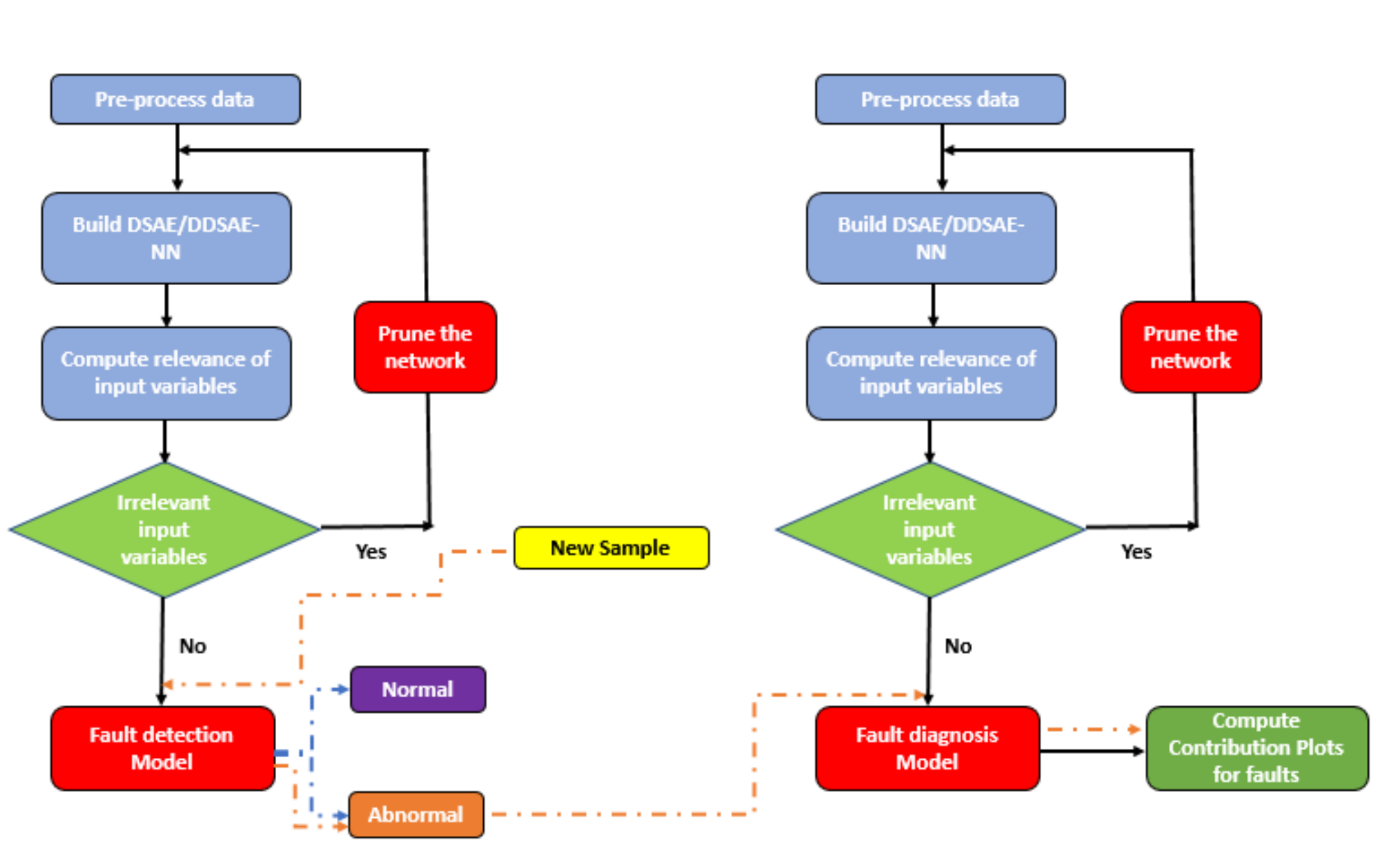}
    \caption{Flowchart for fault detection and diagnosis based on explainable DNN }
    \label{methodology}
\end{figure}

The proposed methodology for FDD is schematically described in Figure \ref{methodology} and it is summarized by the following steps.

\begin{enumerate}
    \item Pre-process the input data. $\textbf{X}_{\text{raw}} \rightarrow \textbf{X}^{l}$.
    \vspace{1em}
    \item Build a DSAE-NN that maps the input vectors into the latent feature space by using a DSAE model structure. The goal is to extract discriminative features that capture the latent manifold in the input data that are most correlated with the output classes. The parameters are optimized using a combination of the reconstruction error, $L_2$ regularization error and binary softmax cross-entropy error of the input data (refer Equation \ref{DSAE_objective_function1}).
    \vspace{1em}
    \item Select the best performing model architecture (number of layers and nodes) along with the set of hyper-parameters that include learning rate, batch size, and all other weighting ($\lambda_1,\lambda_2,\lambda_3$ and $\delta$) parameters using the validation dataset. \big($\textbf{X}^{v},\textbf{y}^{v}$\big).
    \vspace{1em}
    \item Evaluate the classification accuracy using the chosen trained DSAE model in Step 3 for testing dataset \ \big($\textbf{X}^{t},\textbf{y}^{t}$\big). If the model in testing is satisfactory, the model will be used for further analysis; if unsatisfactory, return to Step 3 to redesign the DSAE-NN model.
    \vspace{1em}
    \item Compute input relevances using the LRP method on the test dataset and discard irrelevant input features from the dataset $\textbf{X}^{l}, \textbf{X}^{v}$ and $\textbf{X}^{t}$.
    \vspace{1em}
    \item Repeat steps 3,4 and 5 until relevances of all input variables are above the threshold and no improvement over the validation accuracy is achieved for xDSAE NN.
    \vspace{1em}
    \item Build xDDSAE model using the reduced input dataset $\textbf{X}^{l}_r$ computed in Step 6 along with augmenting $l$ lagged variables.
    \vspace{1em}
    \item Select the best performing model architecture (number of layers and nodes) using the validation dataset \big($\textbf{X}^{v}_r,\textbf{y}^{v}_r$\big).
    \vspace{1em}
    \item Repeat steps 3,4 and 5 until relevances of all input variables are above the threshold and no improvement of the validation accuracy is achieved.
    \vspace{1em}
    \item For online process monitoring: When a new data vector $\textbf{X}_{new}$ becomes available, import it into the model after normalizing to determine whether the current state of the process is in normal or abnormal operating region and to determine the type of fault that is responsible for the deviation.
\end{enumerate}

\begin{figure}[]
\begin{center}
       \includegraphics[width = 0.75\textwidth, height = 24em]{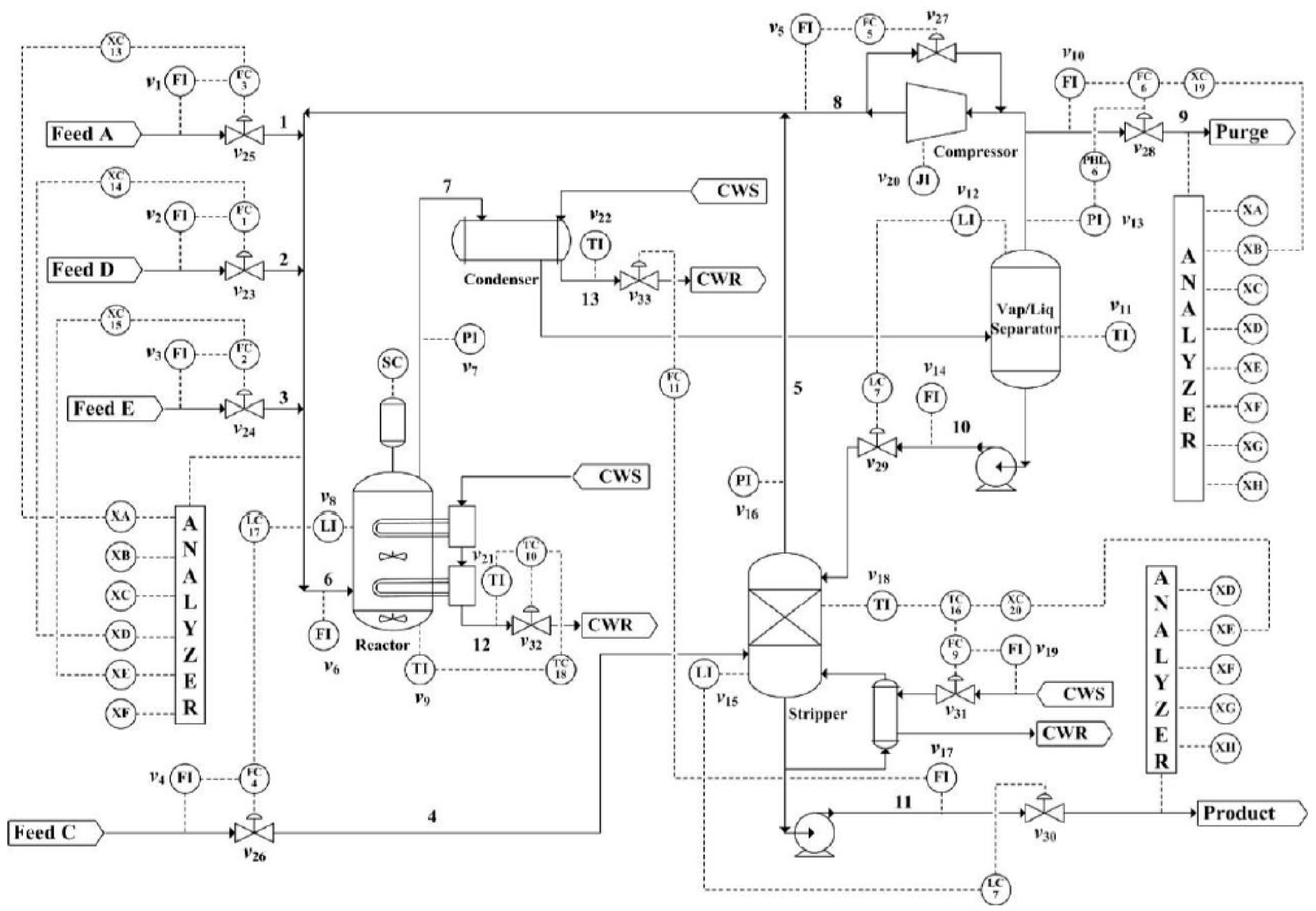}
	\caption{Schematic: Tennessee Eastman plant process (Downs and Vogel, 1993)}
	\label{TEP_schematic} 
    \end{center}
\end{figure}

\begin{table}[width=\linewidth,cols=3]
\centering
\caption{Measured and manipulated variables (from Downs and Vogel, 1993)}
\label{measured_outputs}
\resizebox{\textwidth}{!}{%
\begin{tabular}{@{}lcclcc@{}}
\toprule
\multicolumn{1}{l}{Variable Name} & \multicolumn{1}{c}{Variable Number} & Units & \multicolumn{1}{l}{Variable Name} & \multicolumn{1}{c}{Variable Number} & Units\\ \midrule
A feed (stream 1) & XMEAS (1) & kscmh & Reactor cooling water outlet temperature & XMEAS (21) & $^\circ$ C\\
D feed (stream 2) & XMEAS (2) & kg h$^{-1}$ & Separator cooling water outlet temperature & XMEAS (22) & $^\circ$C \\
E feed (stream 3) & XMEAS (3) & kg h$^{-1}$ & Feed \%A &  XMEAS(23) &  mol\% \\ 
A and C feed (stream 4) & XMEAS (4) & kscmh & Feed \%B &  XMEAS(24) &  mol\% \\ 
Recycle flow (stream 8) & XMEAS (5) & kscmh & Feed \%C &  XMEAS(25) &  mol\% \\ 
Reactor feed rate (stream 6) & XMEAS (6) & kscmh & Feed \%D &  XMEAS(26) &  mol\% \\ 
Reactor pressure & XMEAS (7) & kPa guage & Feed \%E &  XMEAS(27) &  mol\% \\
Reactor level & XMEAS (8) & \% & Feed \%F&  XMEAS(28) &  mol\% \\
Reactor temperature & XMEAS (9) & $^\circ$C &  Purge \%A&  XMEAS(29) &  mol\%\\ 
Purge rate (stream 9) & XMEAS (10) & kscmh & Purge \%B &  XMEAS(30) &  mol\%\\
Product separator temperature & XMEAS (11) & $^\circ$C & Purge \%C &  XMEAS(31) &  mol\%\\
Product separator level & XMEAS (12) & \% & Purge \%D &  XMEAS(32) &  mol\%\\
Product separator pressure & XMEAS (13) & kPa guage & Purge \%E &  XMEAS(33) &  mol\%\\
Product separator underflow (stream 10) & XMEAS (14) & m$^3$ h$^{-1}$ & Purge \%F &  XMEAS(34) &  mol\%\\
Stripper level & XMEAS (15) & \% & Purge \%G &  XMEAS(35) &  mol\%\\
Stripper pressure & XMEAS (16) & kPa guage & Purge \%H &  XMEAS(36) &  mol\%\\
Stripper underflow (stream 11) & XMEAS (17) & m$^3$ h$^{-1}$ & Product \%D & XMEAS(37) & mol\%\\
Stripper temperature & XMEAS (18) & $^\circ$C & Product \%E & XMEAS(38) & mol\%\\
Stripper steam flow & XMEAS (19) & kg h$^{-1}$ & Product \%F & XMEAS(39) & mol\%\\
Compressor Work & XMEAS (20) & kW & Product \%G & XMEAS(40) & mol\%\\
D Feed Flow & XMV (1) & kg h$^{-1}$& Product \%H & XMEAS(41) & mol\% \\
E Feed Flow & XMV (2) & kg h$^{-1}$ & 
A Feed Flow & XMV (3) & kscmh\\
A + C Feed Flow & XMV (4) & kscmh & Compressor Recycle Valve & XMV(5) & \%\\
Purge Valve & XMV (6) & \% &
Separator pot liquid flow & XMV (7) & m$^{3}$h$^{-1}$\\
Stripper liquid product flow & XMV (8) & m$^{3}$h$^{-1}$ & Stripper Steam Valve & XMV (9) & \%\\
Reactor cooling water flow & XMV (10) & m$^{3}$h$^{-1}$ &
Condenser cooling water flow & XMV (11) & m$^{3}$h$^{-1}$\\
\bottomrule
\end{tabular}}
\end{table}

\begin{table}[width=\linewidth,cols=3]
\centering
\caption{Process Faults for classification in TE Process}
\label{process faults}
\resizebox{0.7\textwidth}{!}{%
\begin{tabular}{@{}llc@{}}
\toprule
\multicolumn{1}{l}{Fault} & \multicolumn{1}{c}{Description} & Type \\ \midrule
IDV(1)	& A/C feed ratio, B composition constant (stream 4)	& step\\
IDV(2)	& B composition, A/C ratio constant (stream 4)	& step\\
IDV(3) & D Feed Temperature & step\\
IDV(4)	& Reactor cooling water inlet temperature	& step\\
IDV(5)	& Condenser cooling water inlet temperature (stream 2) & step\\
IDV(6) &	A feed loss (stream 1) & step\\
IDV(7) & C header pressure loss reduced availability (stream 4)	& step\\
IDV(8) &	A, B, C feed composition (stream 4)	& random variation\\
IDV(9) & D Feed Temperature & random variation\\
IDV(10) & C feed temperature (stream 4)	& random variation\\
IDV(11) & Reactor cooling water inlet temperature & random variation\\
IDV(12)	& Condenser cooling water inlet temperature	 & random variation\\
IDV(13) & Reaction kinetics & 	slow drift\\
IDV(14)	& Reactor cooling water & valve	sticking\\
IDV(15) & Condenser Cooling Water Valve & stiction\\
IDV(16) & Deviations of heat transfer within stripper & random variation\\
IDV(17) & Deviations of heat transfer within reactor & random variation\\
IDV(18) & Deviations of heat transfer within condenser & random variation\\
IDV(19) & Recycle valve of compressor, underflow stripper and steam valve stripper & stiction\\
IDV(20) & unknown & random variation \\
\bottomrule
\end{tabular}}
\end{table}

\section{Case Study: Tennessee Eastman Process}
In this section the proposed methodology is implemented for FDD and the performance is compared with different approaches in the literature on the benchmark Tennessee Eastman Process (TEP). The Tennessee Eastman plant has been used widely for testing several process monitoring and fault detection algorithms \cite{chiang2000fault,lau2013fault,rato2013fault,xie2015hierarchical,ricker1996decentralized,bathelt2015revision,kulkarni2005knowledge,larsson2001self}. The TEP involves different unit operations including a vapor-liquid separator, a reactor, stripper a recycle compressor and a condenser. Four gaseous reactants (A, B, C and D) forms two liquid products streams (G and H) and a by-product (F). A schematic of the Tennessee Eastman Process is illustrated in Figure \ref{TEP_schematic}. \citeauthor{downs1993plant} (\citeyear{downs1993plant})\cite{downs1993plant} reported the original simulator for this process and has been widely used as a benchmark process for control and monitoring studies (simulator available at \url{http://depts.washington.edu/control/LARRY/TE/download.html}). The process simulator involves a total of $52$ measured variables including $22$ process (output) variables, $12$ manipulated variables and $19$ composition measurements. A complete list of output measurements and manipulated variables are presented in Table \ref{measured_outputs}. Additional details about the process model can be found in the original paper \cite{downs1993plant} and descriptions of the different control schemes that have been applied to the simulator can be found in \cite{ricker1996decentralized} and its revised version \cite{bathelt2015revision}. Several data-driven statistical process monitoring approaches have been reported for the detection and diagnosis of disturbances in the Tennessee Eastman simulation. There are $20$ different process disturbances (fault types) in the industrial simulator (shown in Table \ref{process faults}) though only $17$ were used in this work to be consistent with other methods in the literature. Each of these methods has shown different levels of success in detecting and diagnosing the faults considered in the simulations (Table 1).  Several statistical studies have reported faults 3, 9 and 15 as unobservable or difficult to diagnose due to the close similarity in the responses of the noisy measurements used to detect these faults \cite{lau2013fault,shams2010fault,chiang2000fault,du2018fault} and therefore these 3 faults were not considered in the current study.\\

The training data consists of $500$ samples of normal data and $480$ samples for each fault. The testing dataset has $960$ samples for both faulty and normal operation data. For the faulty testing dataset, the fault is introduced at $160$ time-sample. A part of the training data $\{\textbf{X}^{l},\textbf{y}^{l}\}$ is used as the validation dataset $\{\textbf{X}^{v},\textbf{y}^{v}\}$ for tuning the hyper-parameters (learning rate, weights: $\lambda, \lambda_{1}, ~\lambda_{2}, ~\lambda_{3}$ and $\delta$, number of epochs,layers an nodes in each layer) for both DSAE/ xDSAE and DDSAE/ xDDSAE DNNs for all iterations. The network architecture and test accuracies for fault detection and fault classification are presented in Table \ref{architecture_fault_detection} and \ref{architecture_fault_diagnosis} respectively. After applying the LRP procedure to the static fault detection model, it is found that only 24 out of 52 variables are the most important for obtaining the highest testing accuracy for detecting the correct state of the process plant. After every iteration of the input pruning-relevance (LRP) based procedure it is shown in Table \ref{architecture_fault_detection} that the removal of irrelevant input variables results in successive improvement of fault class separability. 
To account for the dynamic information after identifying the 24 most relevant process variables, the reduced input data matrix $\{\textbf{X}^{l}_r\}$ is stacked with lagged time stamps and an DDSAE NN model is retrained. The best fault detection test accuracy of $96.43\%$ is achieved by stacking two previous time-stamp process values. The fault detection rates for all the faults are shown in Table
\ref{comparison-detection}. These results are compared in the same Table \ref{comparison-detection} with several methods as follows:  PCA \cite{lv2016fault}, DPCA\cite{lv2016fault}, ICA\cite{hsu2010novel}, Convolutional NN (CNN) \cite{8926924}, Deep Stacked Network (DSN) \cite{chadha2017comparison}, Stacked Autoencoder (SAE) \cite{chadha2017comparison}, Generative Adversarial Network (GAN) \cite{spyridon2018generative} and One-Class SVM (OCSVM) \cite{spyridon2018generative}. It can been seen from Table \ref{comparison-detection} that the proposed method outperformed the linear multivariate methods and other DL based methods for most fault modes. For example, for PCA with 15 principal components, the average fault detection rates are 61.77\% and 74.72\% using $T^2$ and $Q$ statistic respectively. Since the principal components extracted using PCA captures static correlations between variables, DPCA (Dynamic PCA) is used to account for temporal correlations (both auto-correlations and cross-correlations) in the data. Since DPCA is only an input data compression technique, it must be combined with a classification model for the purpose of fault detection. Accordingly, the output features from the DPCA model are fed into an SVM model that is used for final classification. The effect of increasing the number of lagged variables in the dataset is also investigated following the hypothesis that increasing the time horizon will enhance classification accuracy. It can be seen in Table 3 that the increasing number of lags and simultaneous pruning improves the classification accuracy. 
The average detection rate obtained was 72.35\%. ICA \cite{hsu2010novel} based monitoring scheme was found to perform better than both PCA and DPCA based methods with an averaged accuracy of approximately 90\%. In addition to the comparison to linear methods, the proposed methodology was also compared with different DNN architectures such as CNN\cite{chadha2017comparison}, DSN \cite{chadha2017comparison}, SAE-NN (results reported in \citeauthor{chadha2017comparison},\citeyear{chadha2017comparison}) and GAN\cite{spyridon2018generative}, OCSVM (results reported in \citeauthor{spyridon2018generative},\citeyear{spyridon2018generative}) reported previously. It can be seen that the proposed method also outperforms these DNN based methodologies. 
Also, the false alarm rate (FAR) i.e. normal samples miss-classified as faulty is $1.46\%$ which is the lowest as compared to all the other methods.\\

\begin{table}[width=\linewidth,cols=4,pos=h]
\caption{Network Architecture and iterations for fault detection methodology}
\label{architecture_fault_detection}
\begin{threeparttable}
\begin{tabular}{@{}cccc@{}}
\toprule
Iteration & \begin{tabular}[c]{@{}c@{}}Network Type\\ DSAE/ DDSAE\end{tabular} & Architecture & \begin{tabular}[c]{@{}c@{}}Averaged Test Classification Accuracy\\ (FDR)\end{tabular} \\ \midrule
1 & DSAE & $52-5-10^{*}-10-5-52$ & 91.55\% \\
2 & xDSAE & $30-5-10^{*}-10-5-30$ & 93\% \\
3 & xDSAE & $24-7-6^{*}-6-7-30$ & 93.23\% \\ \toprule
1 & DDSAE (lag1) & $48-4-6^{*}-6-4-48$ & 93.96\% \\
2 & xDDSAE (lag1) & $46-5-7^{*}-7-5-46$ & 95.52\% \\
3 & xDDSAE (lag1) & $41-5-10^{*}-10-5-41$ & 95.63\% \\
4 & xDDSAE (lag1) & $40-5-10^{*}-10-2-40$ & 95.85\% \\ \toprule
1 & DDSAE (lag2) & $72-4-10^{*}-10-4-72$ & 93.5\% \\
2 & xDDSAE (lag2) & $70-2-10^{*}-10-2-70$ & 93.53\% \\
3 & xDDSAE (lag2) & $54-4-10^{*}-10-4-54$ & 95.44\% \\
4 & xDDSAE (lag2) & $50-6-12^{*}-12-6-50$ & 96.43\% \\ \bottomrule
\end{tabular}
\begin{tablenotes}
  \item[*] A dense layer is present where the number of input nodes are shown with an asterisk and the number of output nodes are 2 (equal to the number of classes).
  \end{tablenotes}
  \end{threeparttable}
\end{table}

\begin{figure}[]
    \centering
    \includegraphics[width = \textwidth]{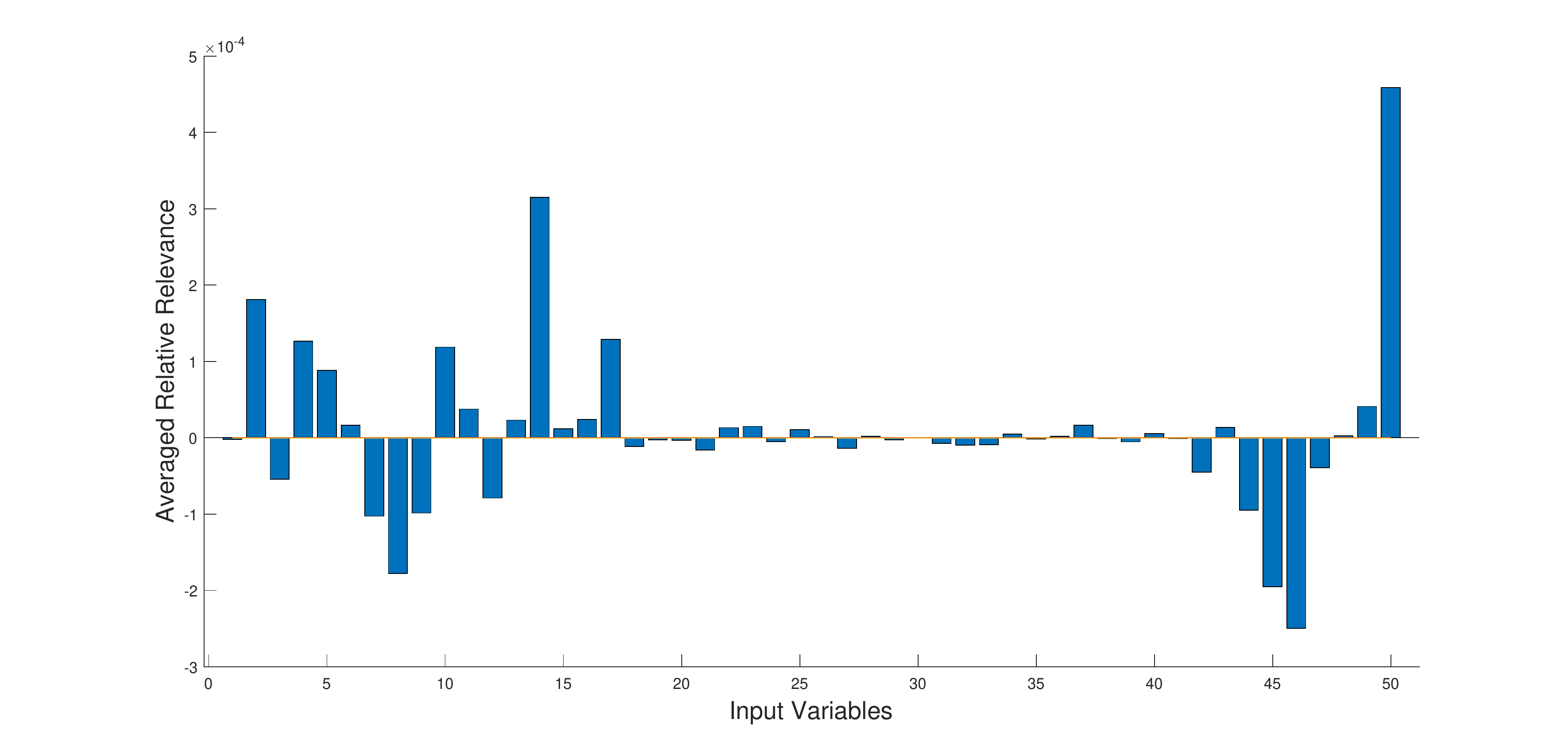}
    \caption{Final Iteration: Averaged Relative Relevance Plot for Fault Detection (DDSAE Model with 2 lagged input variables as $\textbf{X}^{l}$)}
    \label{fig:my_label}
\end{figure}

\begin{table}[]
\caption{Network Architecture and iterations for fault diagnosis methodology}
\label{architecture_fault_diagnosis}
\begin{threeparttable}
\begin{tabular}{@{}cccc@{}}
\toprule
Iteration & \begin{tabular}[c]{@{}c@{}}Network Type\\ DSAE/ DDSAE\end{tabular} & Architecture & \begin{tabular}[c]{@{}c@{}}Averaged Test Classification\\ Accuracy (FDR)\end{tabular} \\ \midrule
1 & DSAE & $52-25-20-20^{*}-20-20-25-52$ & 81.90\% \\
2 & xDSAE & $45-10-10-20^{*}-20-10-10-45$ & 82.60\% \\
3 & xDSAE & $33-21-20-20^{*}-20-20-21-33$ & 83.15\% \\ \toprule
1 & DDSAE (5 lags) & $198-14-10-30^{*}-30-10-14-198$ & 83.41\% \\
2 & xDDSAE (5 lags) & $159-24-10-30^{*}-30-10-24-159$ & 85.14\% \\
3 & xDDSAE (5 lags) &  $155-24-8-30^{*}-30-8-24-155$ & 85.87\% \\
4 & xDDSAE (5 lags) & $140-30-20-17^{*}-17-20-30-140$ & 86.91\% \\ \toprule
1 & DDSAE (10 lags) & $363-14-20-30^{*}-30-20-14-363$ & 83.08\% \\
2 & xDDSAE (10 lags) & $317-18-15-30^{*}-30-15-18-317$ & 85.04\% \\
3 & xDDSAE (10 lags) & $293-24-18-30^{*}-30-18-24-293$ & 85.51\% \\
4 & xDDSAE (10 lags) & $259-28-18-30^{*}-30-18-28-259$ & 87.07\% \\
5 & xDDSAE (10 lags) & $244-34-20-30^{*}-30-20-34-244$ & 87.86\% \\
6 & xDDSAE (10 lags) & $235-38-21-30^{*}-30-21-38-235$ & 88.41\% \\ \bottomrule
\end{tabular}
\begin{tablenotes}
  \item[*] A dense layer is present where the number of input nodes are shown with an asterisk and the number of output nodes are 17 (equal to the number of classes).
  \end{tablenotes}
  \end{threeparttable}
\end{table}

\begin{figure}[]
    \centering
    \includegraphics[width = 0.7\textwidth]{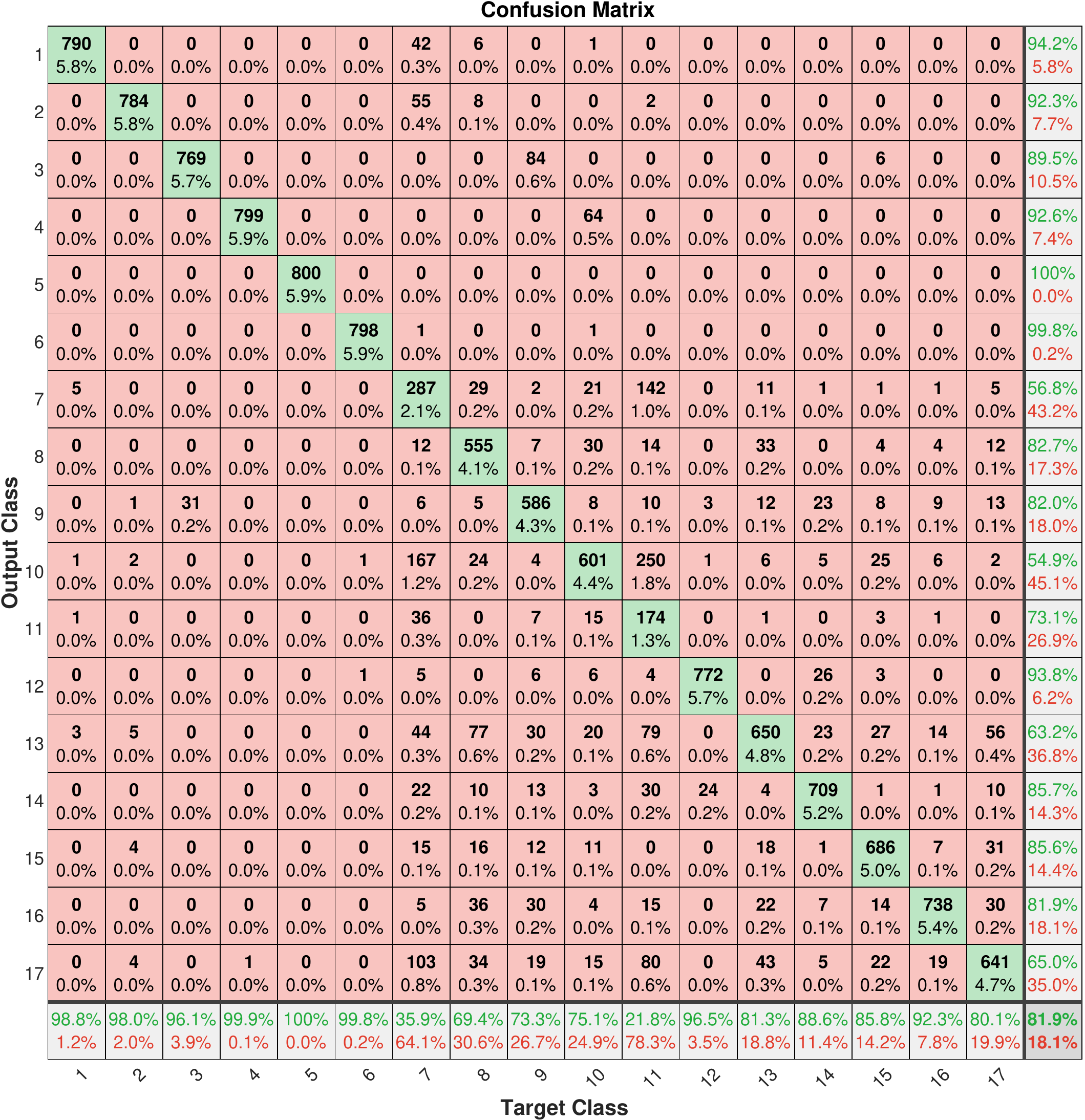}
    \caption{Confusion Matrix for Fault Classification (First Iteration: DSAE Model with 52 input variables)}
    \label{fault_classification_confusion_matrix:initial}
\end{figure}

\begin{figure}[]
    \centering
    \includegraphics[width = 1\textwidth]{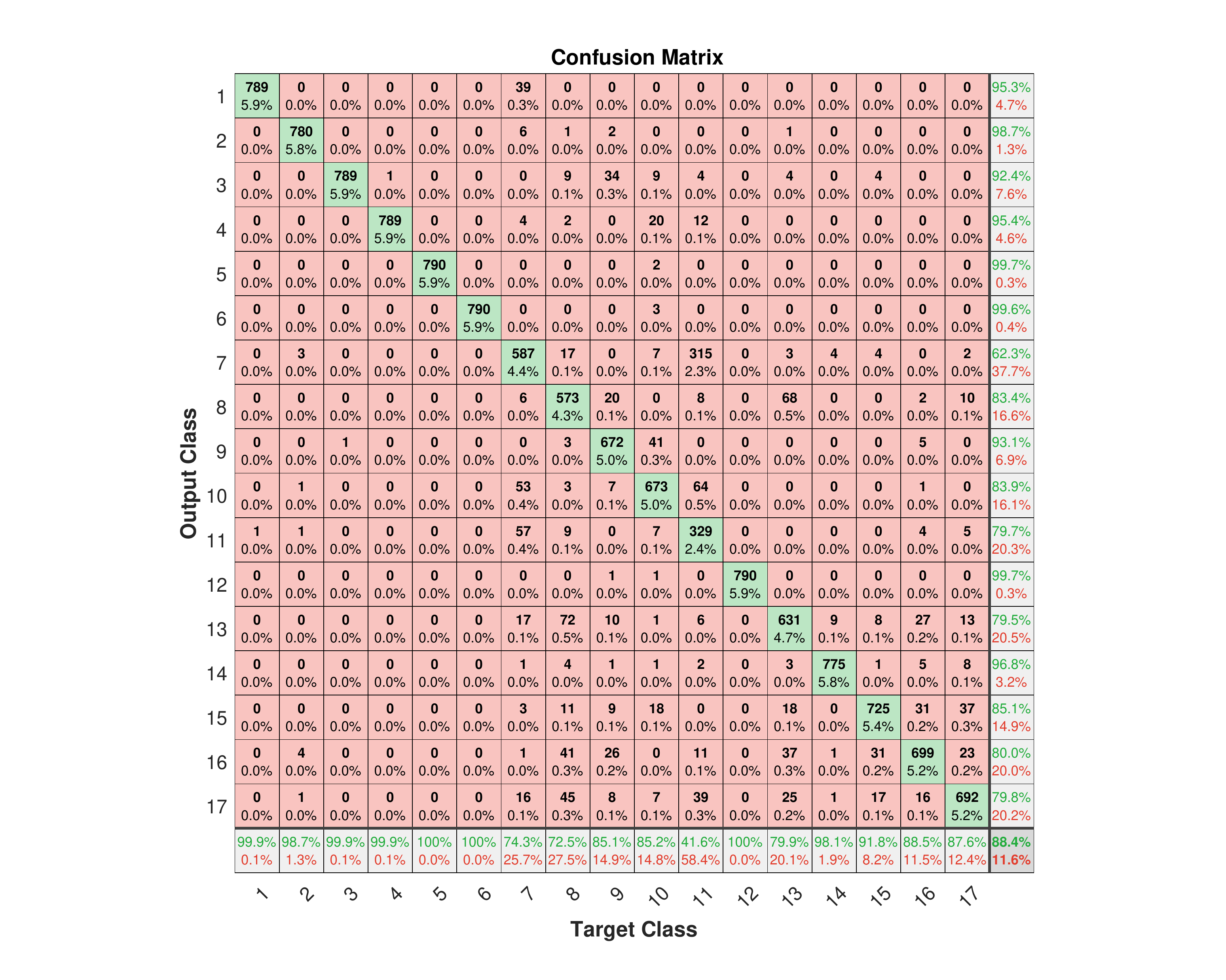}
    \caption{Confusion Matrix for Fault Classification (Final Iteration: DDSAE Model with 10 lagged input variables)}
    \label{fault_classification_confusion_matrix:final}
\end{figure}

\begin{figure}[]
    \centering
    \includegraphics[width = \textwidth]{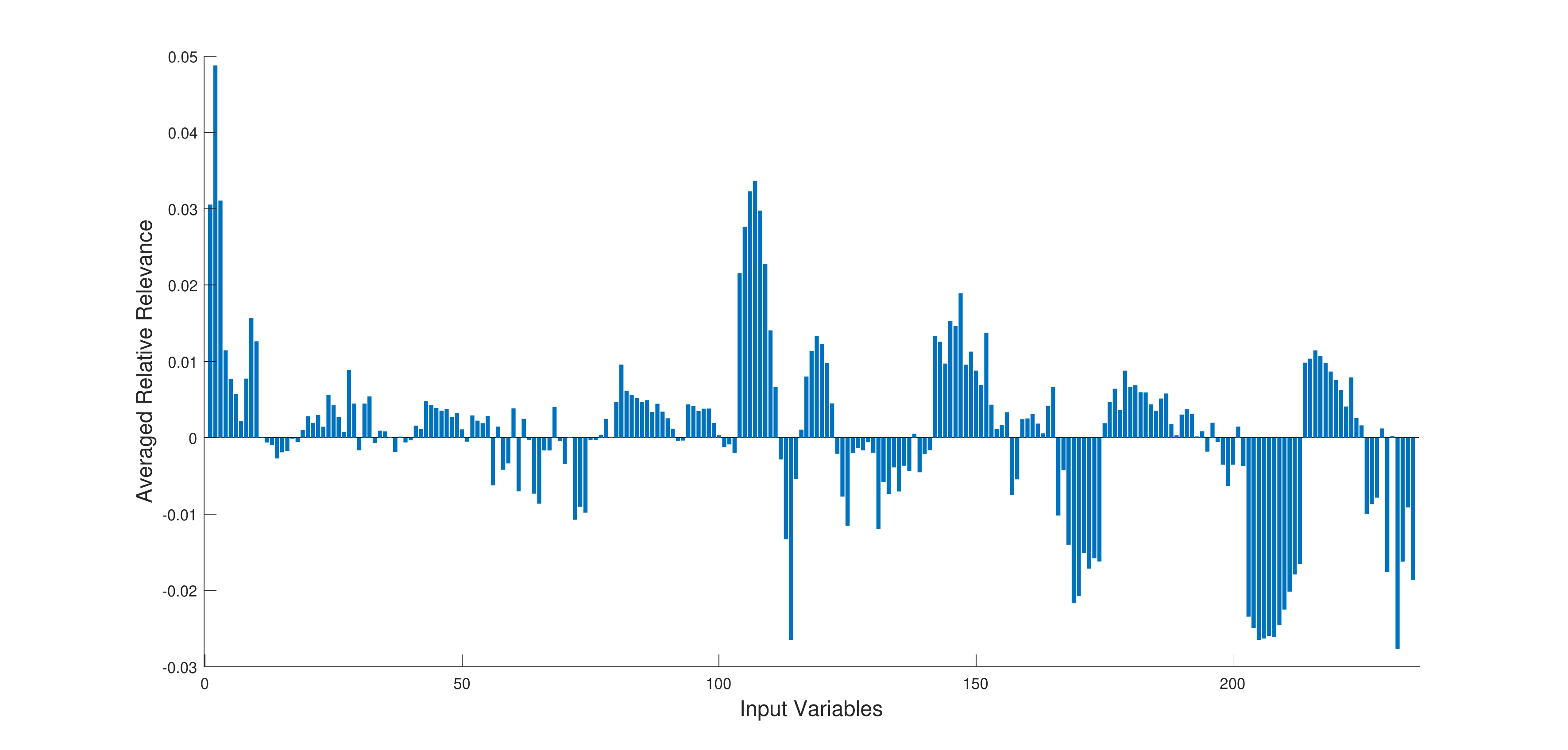}
    \caption{Final Iteration: Averaged Relative Relevance Plot for Fault Diagnosis (DDSAE Model with 10 lagged input variables as $\textbf{X}^{l}$)}
    \label{fault_classification_relevance}
\end{figure}

\begin{figure}[]
    \centering
    \includegraphics[width = \textwidth]{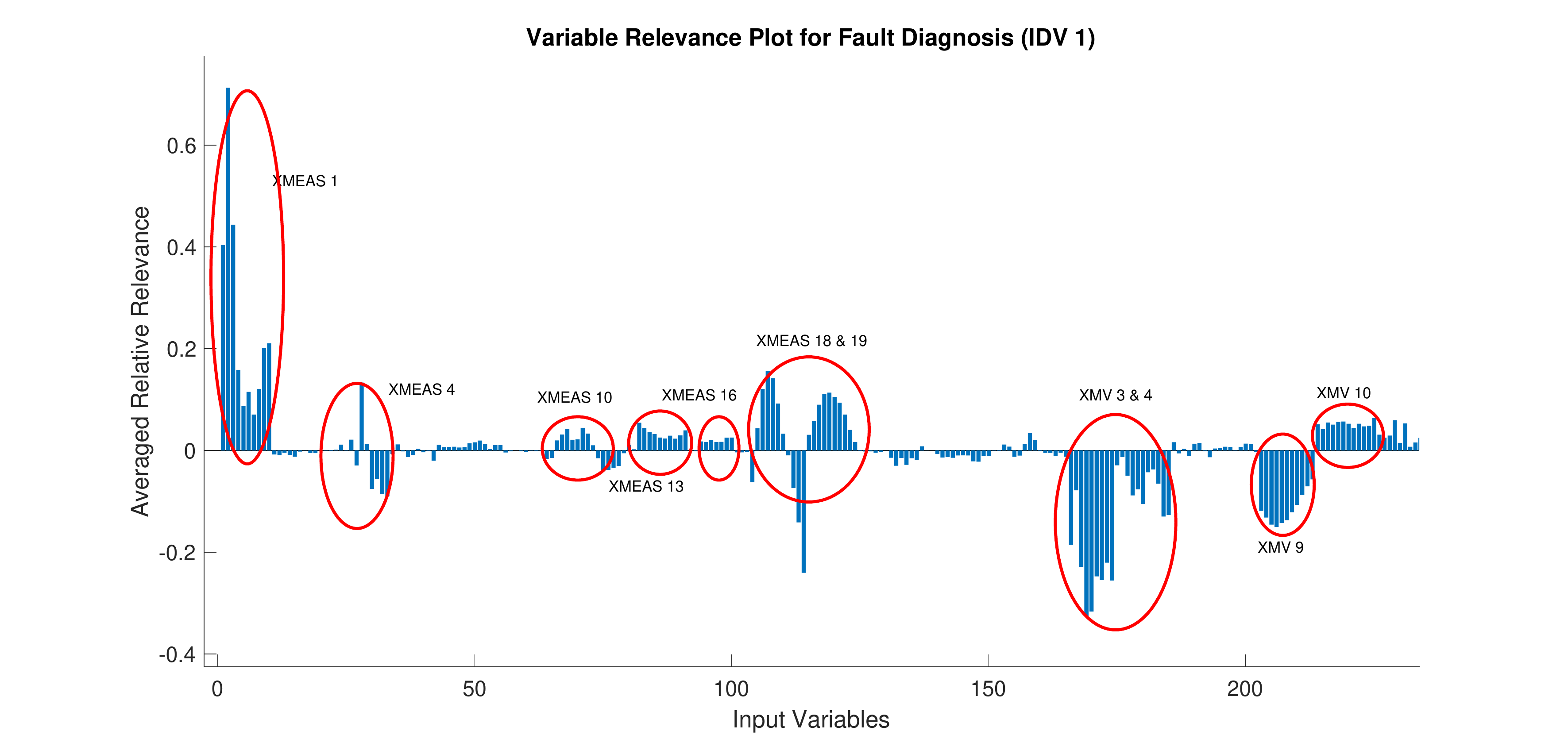}
    \caption{Input variable relevance plot for Fault Diagnosis (IDV 1)}
    \label{relevance for Fault 1}
\end{figure}

\begin{figure}[]
    \centering
    \includegraphics[width = 0.32\textwidth]{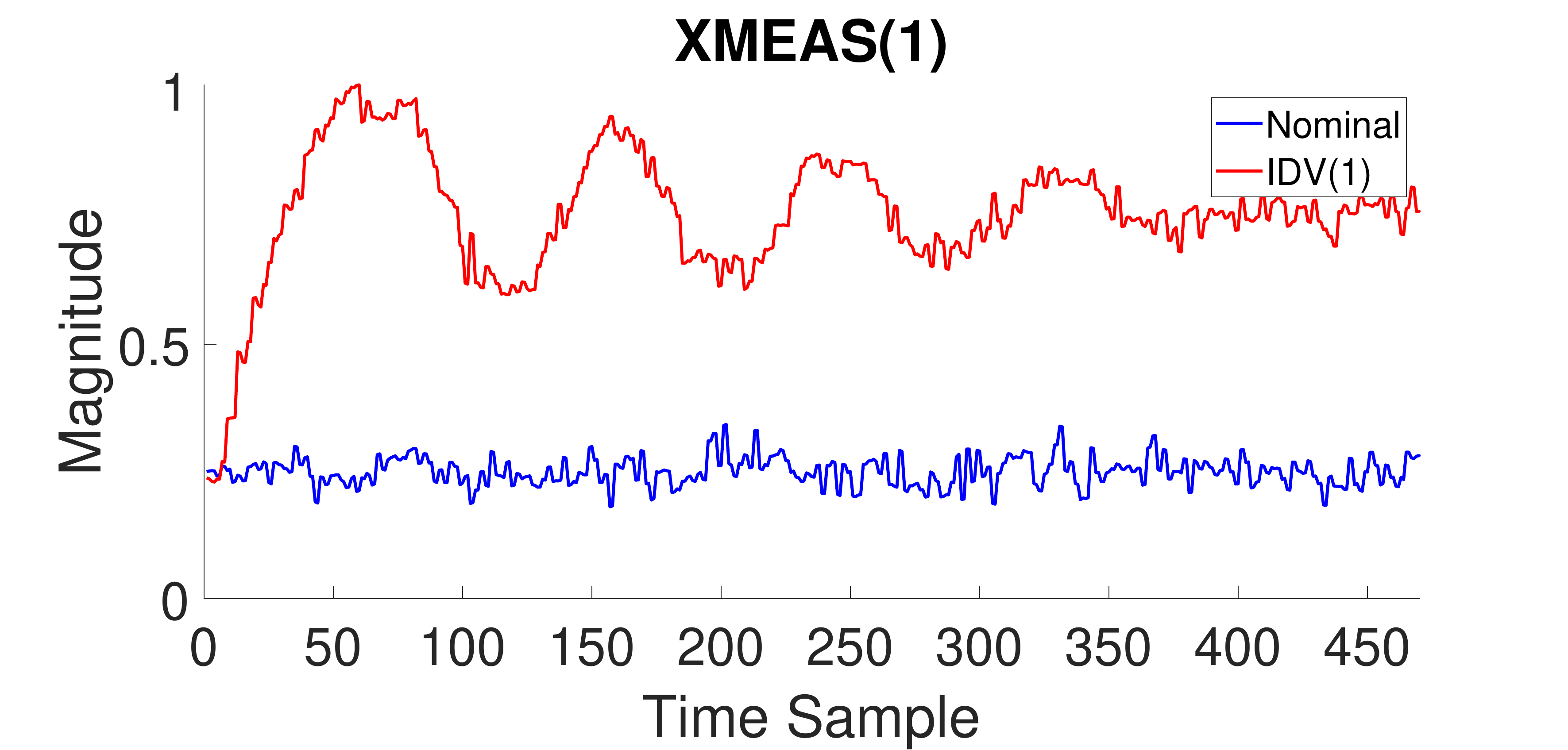}
    \includegraphics[width = 0.32\textwidth]{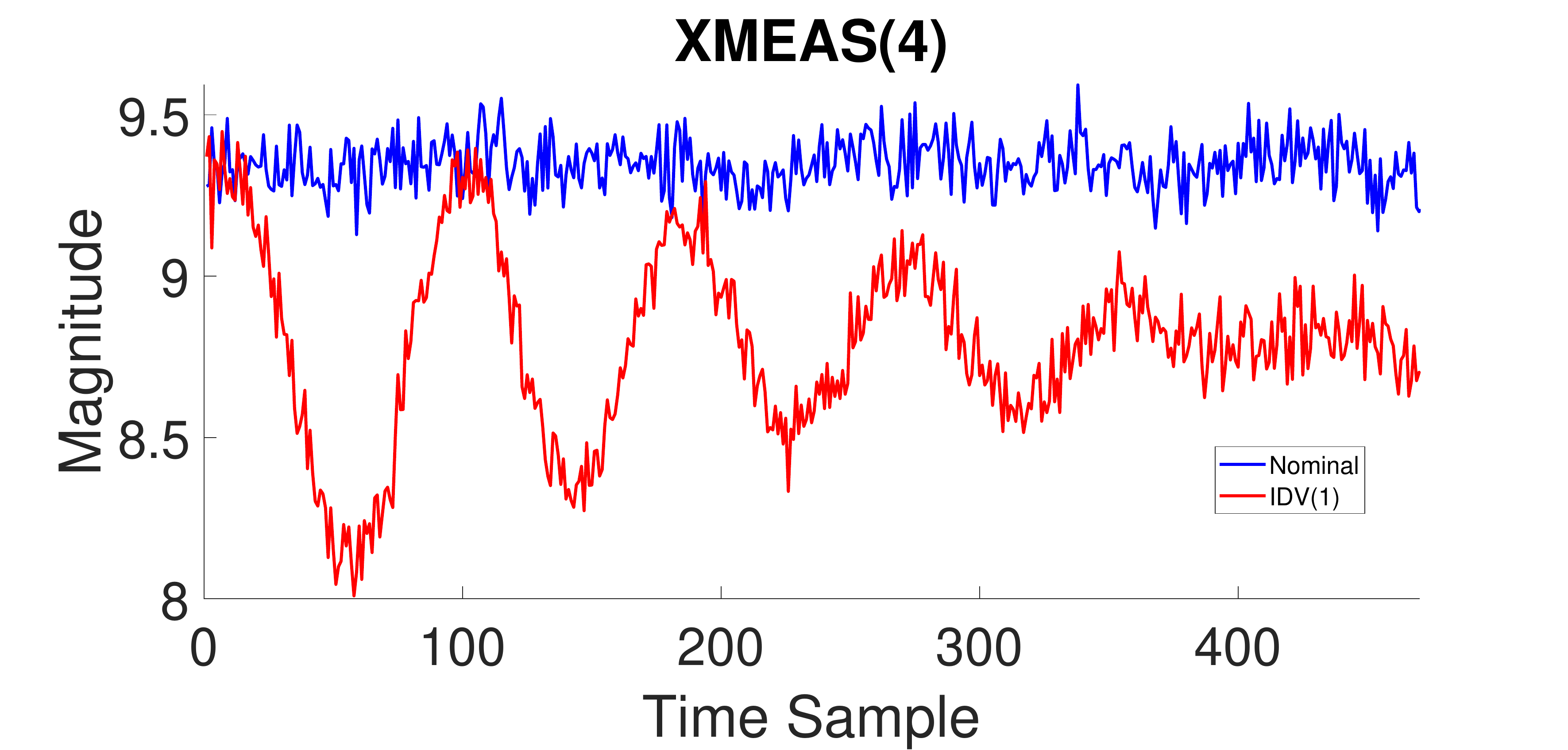}
    \includegraphics[width = 0.32\textwidth]{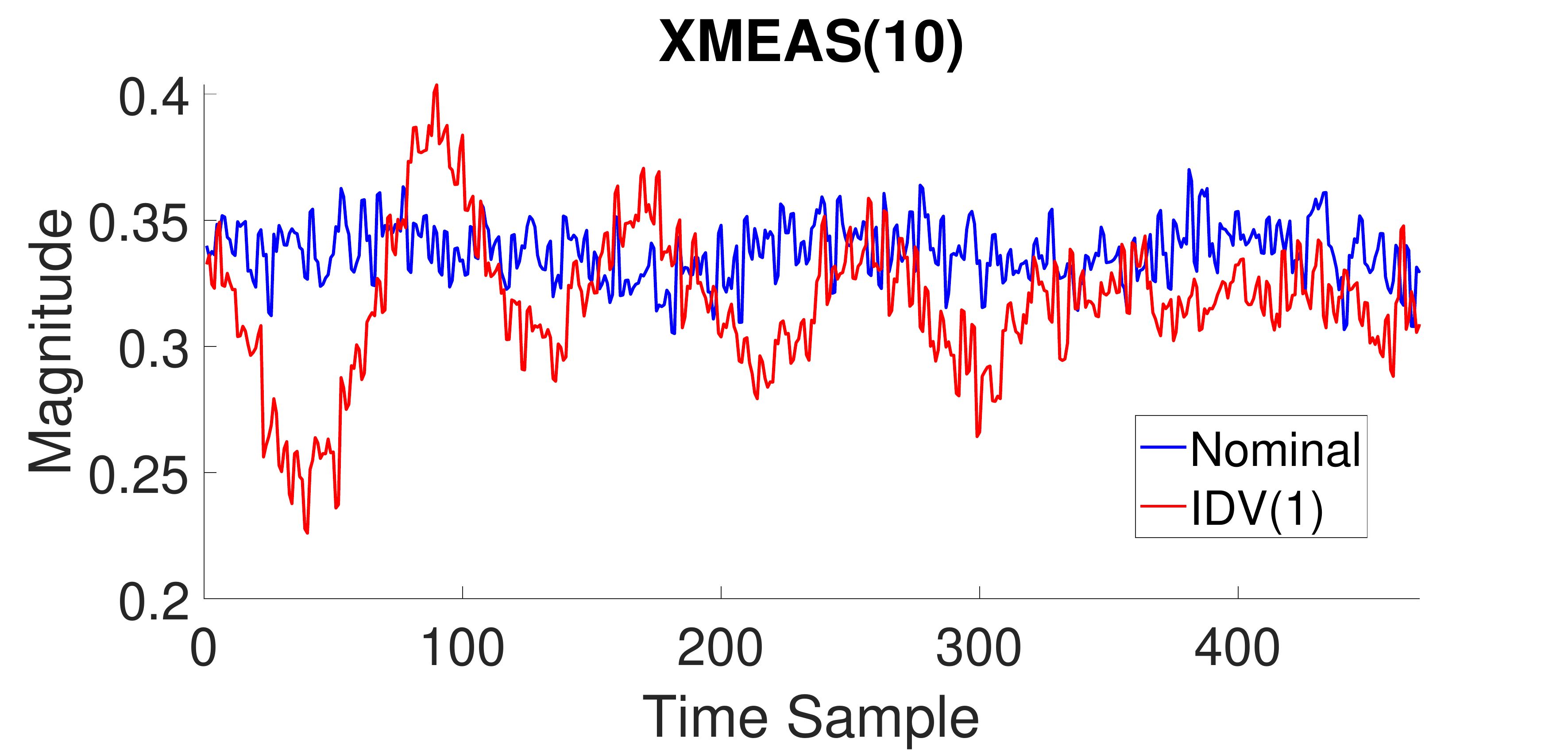}
    \includegraphics[width = 0.32\textwidth]{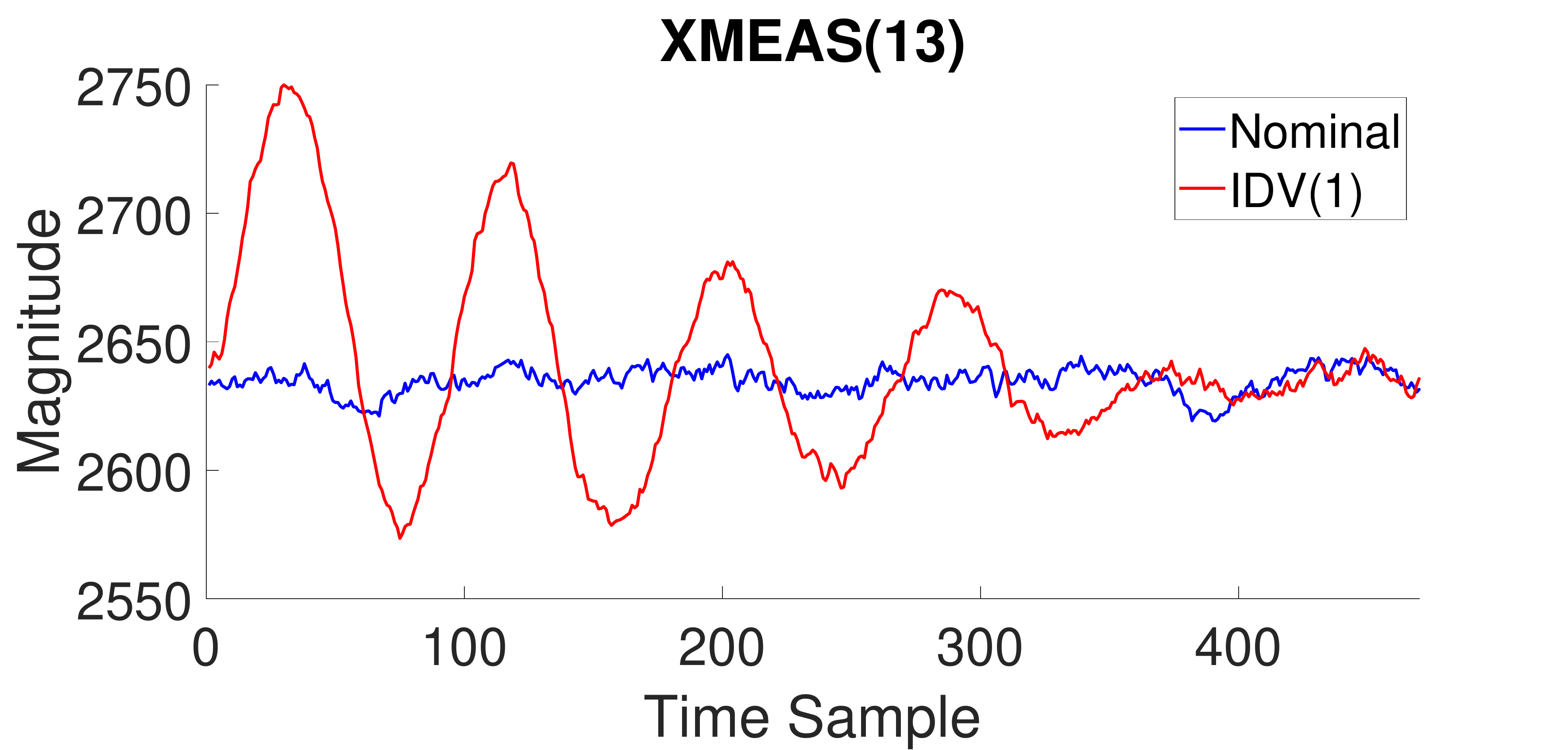}
    \includegraphics[width = 0.32\textwidth]{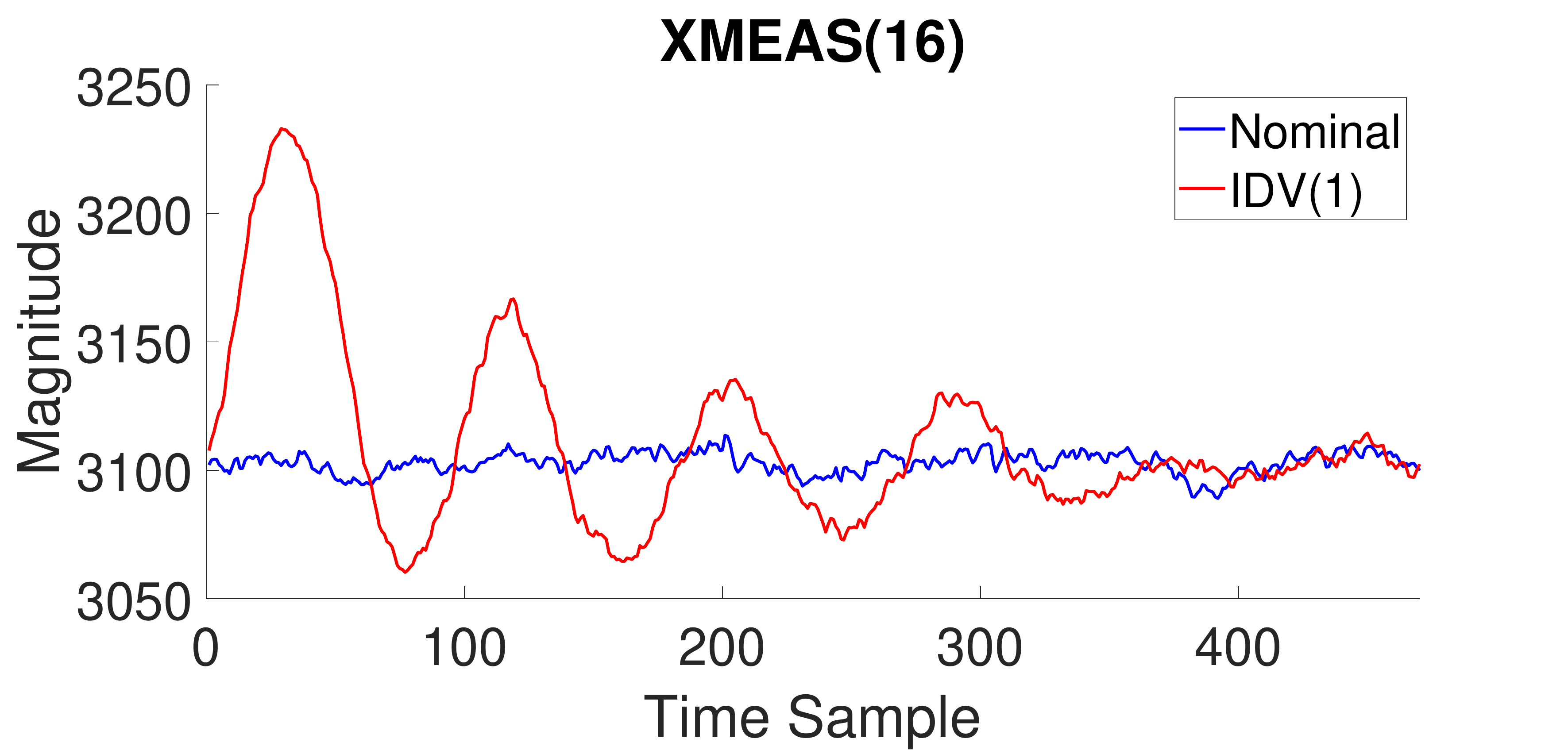}
    \includegraphics[width = 0.32\textwidth]{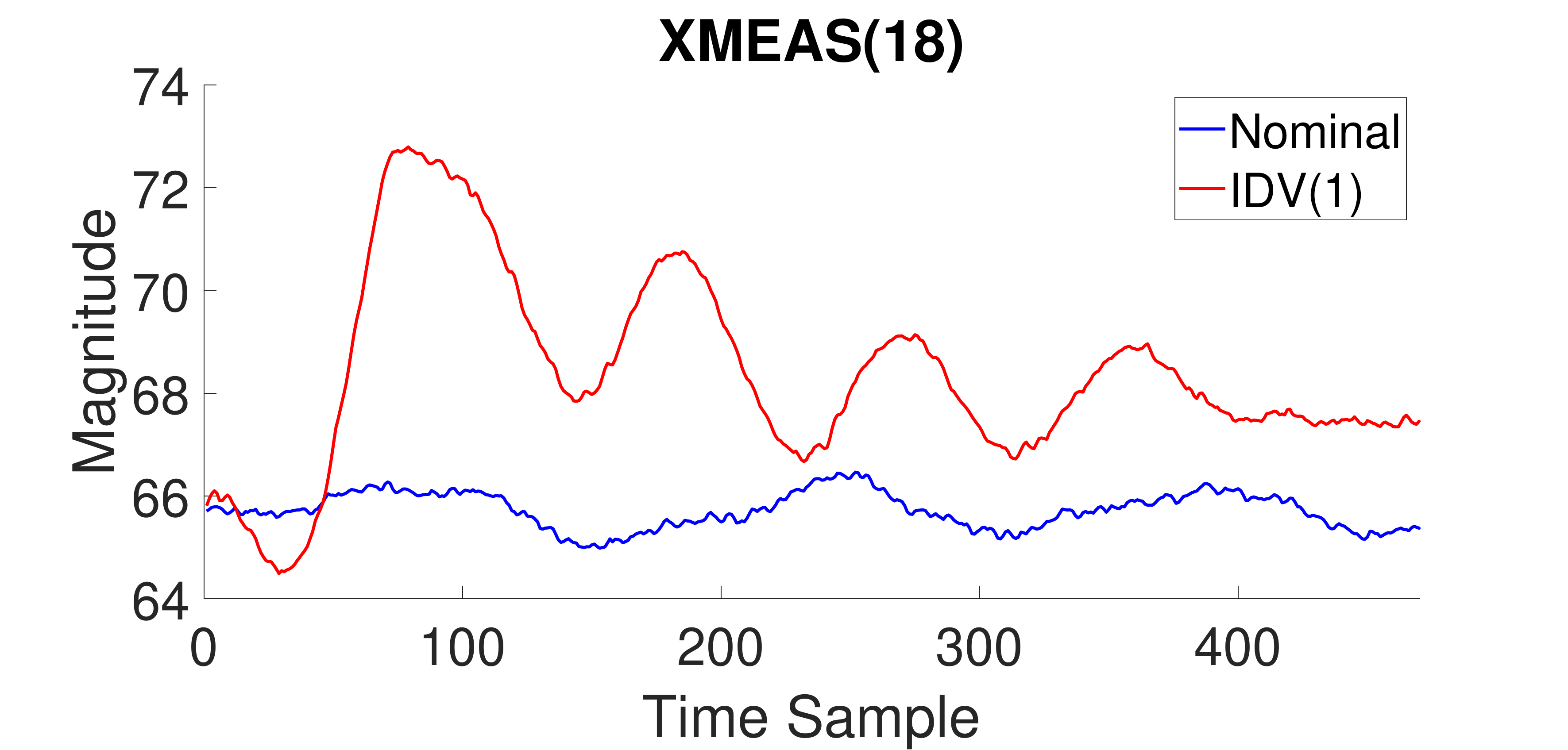}
    \includegraphics[width = 0.32\textwidth]{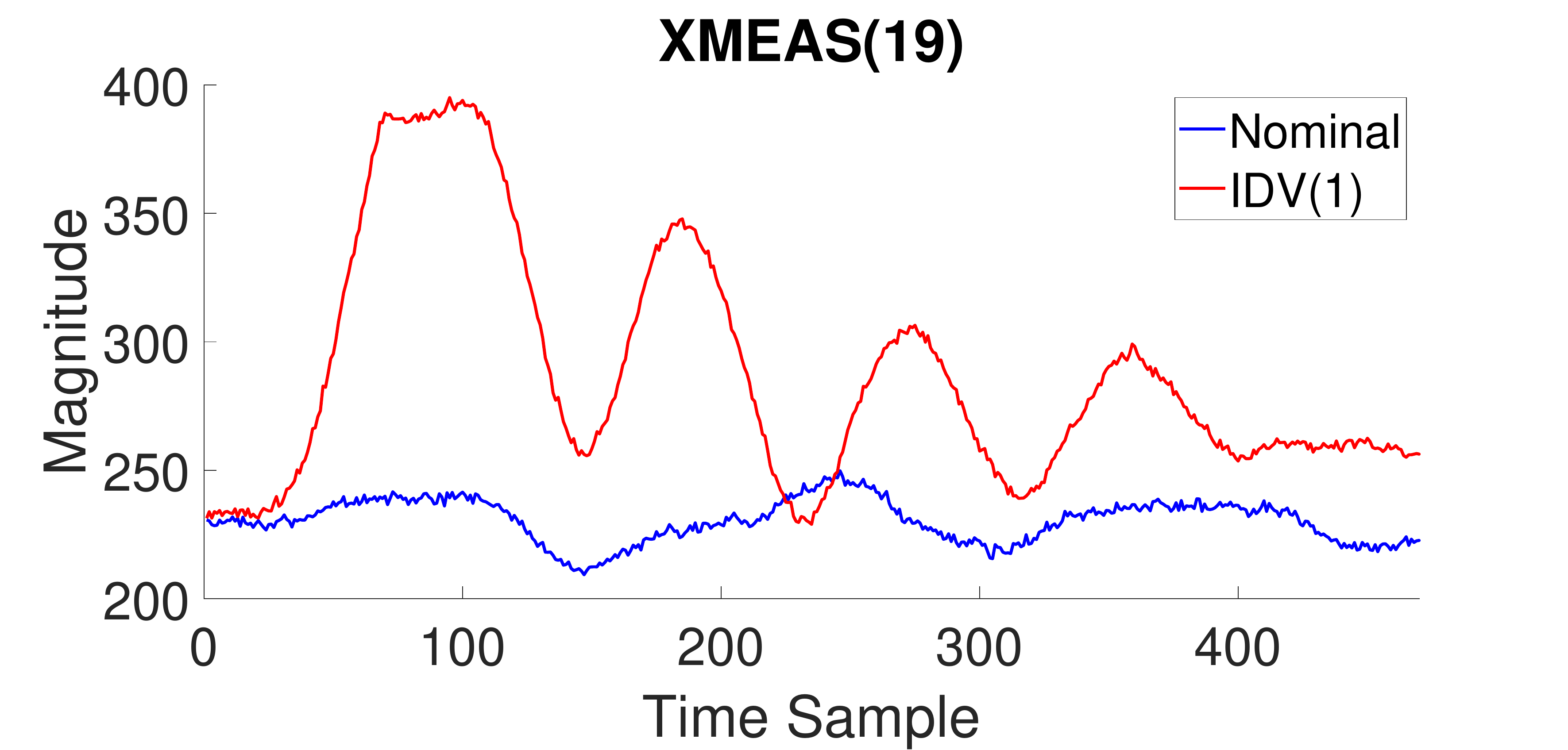}
    \includegraphics[width = 0.32\textwidth]{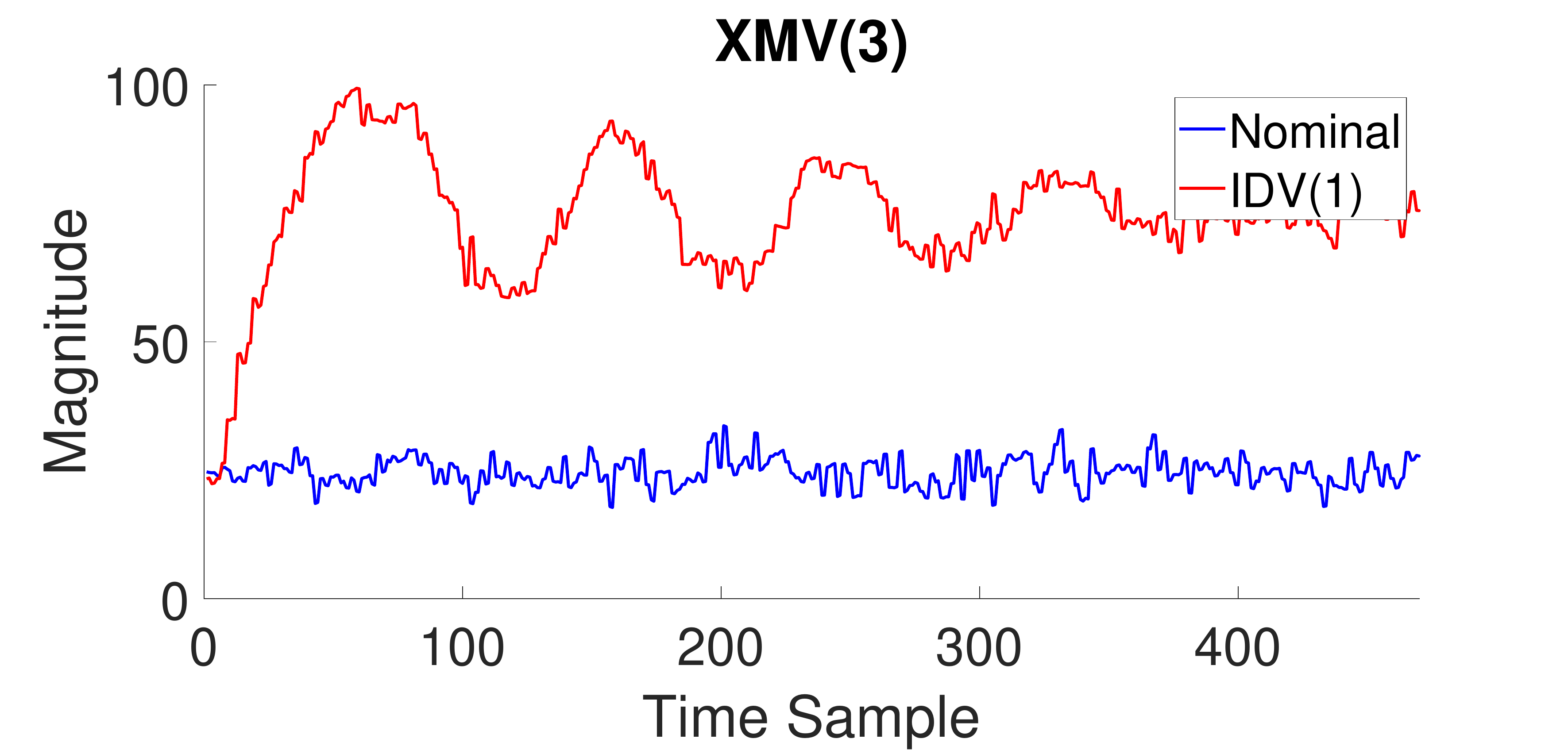}
    \includegraphics[width = 0.32\textwidth]{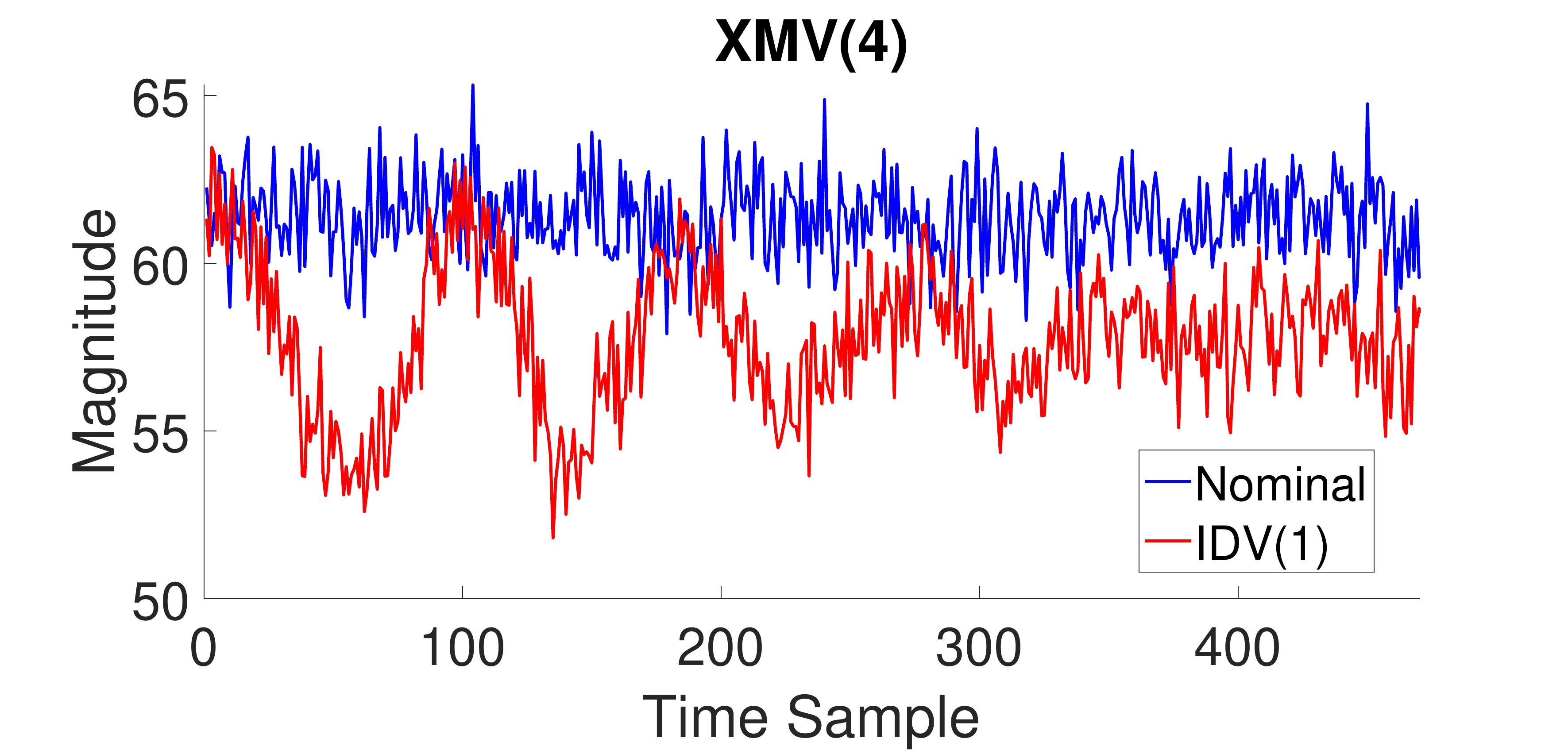}
    \includegraphics[width = 0.32\textwidth]{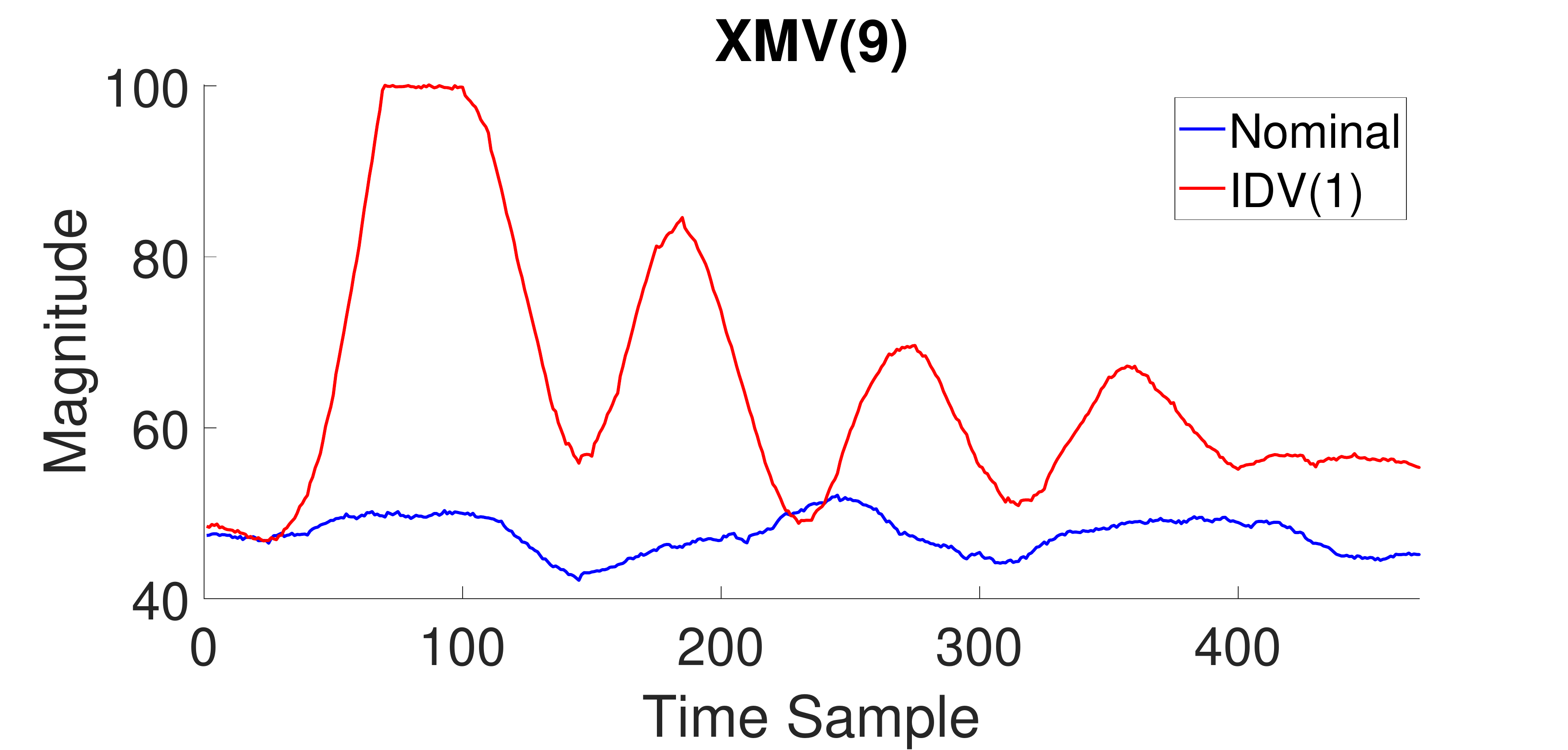}
    \includegraphics[width = 0.32\textwidth]{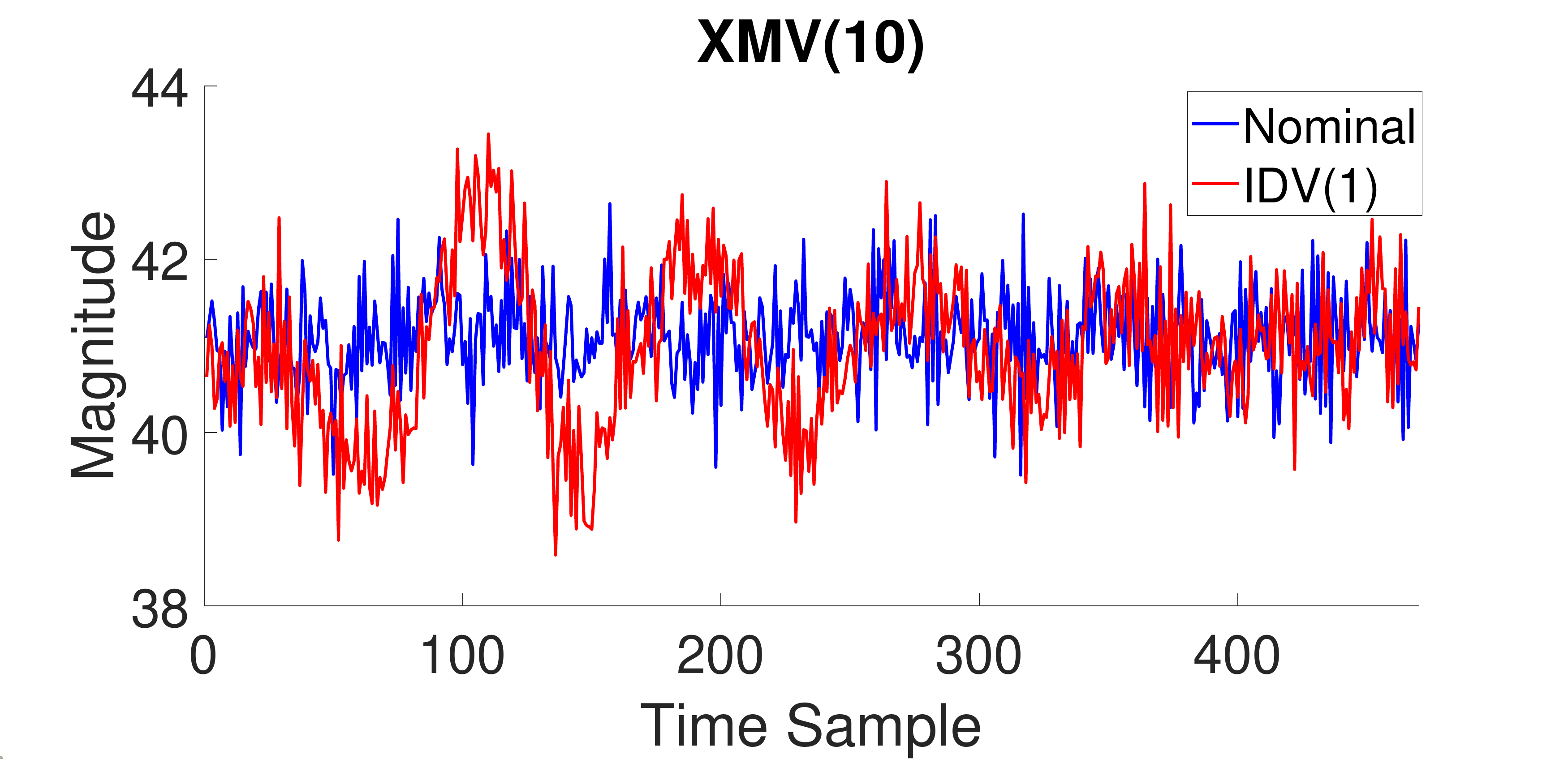}
    \caption{Relevant variables contributing to IDV(1) with nominal and abnormal profiles}
    \label{deviation_nominal_1}
\end{figure}

\begin{figure}[]
    \centering
    \includegraphics[width = \textwidth]{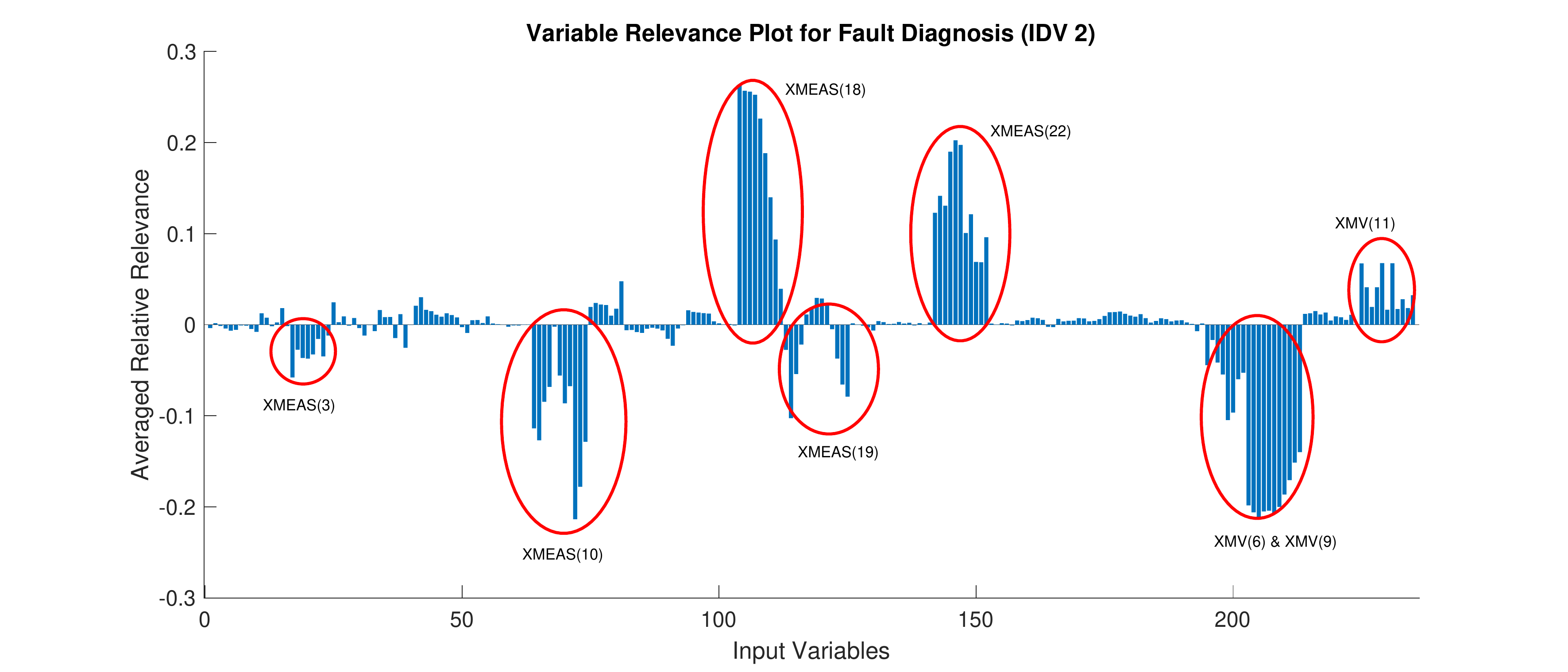}
    \caption{Input variable relevance plot for Fault Diagnosis (IDV 2)}
    \label{relevance for Fault 2}
\end{figure}

\begin{figure}[]
    \centering
    \includegraphics[width = 0.32\textwidth]{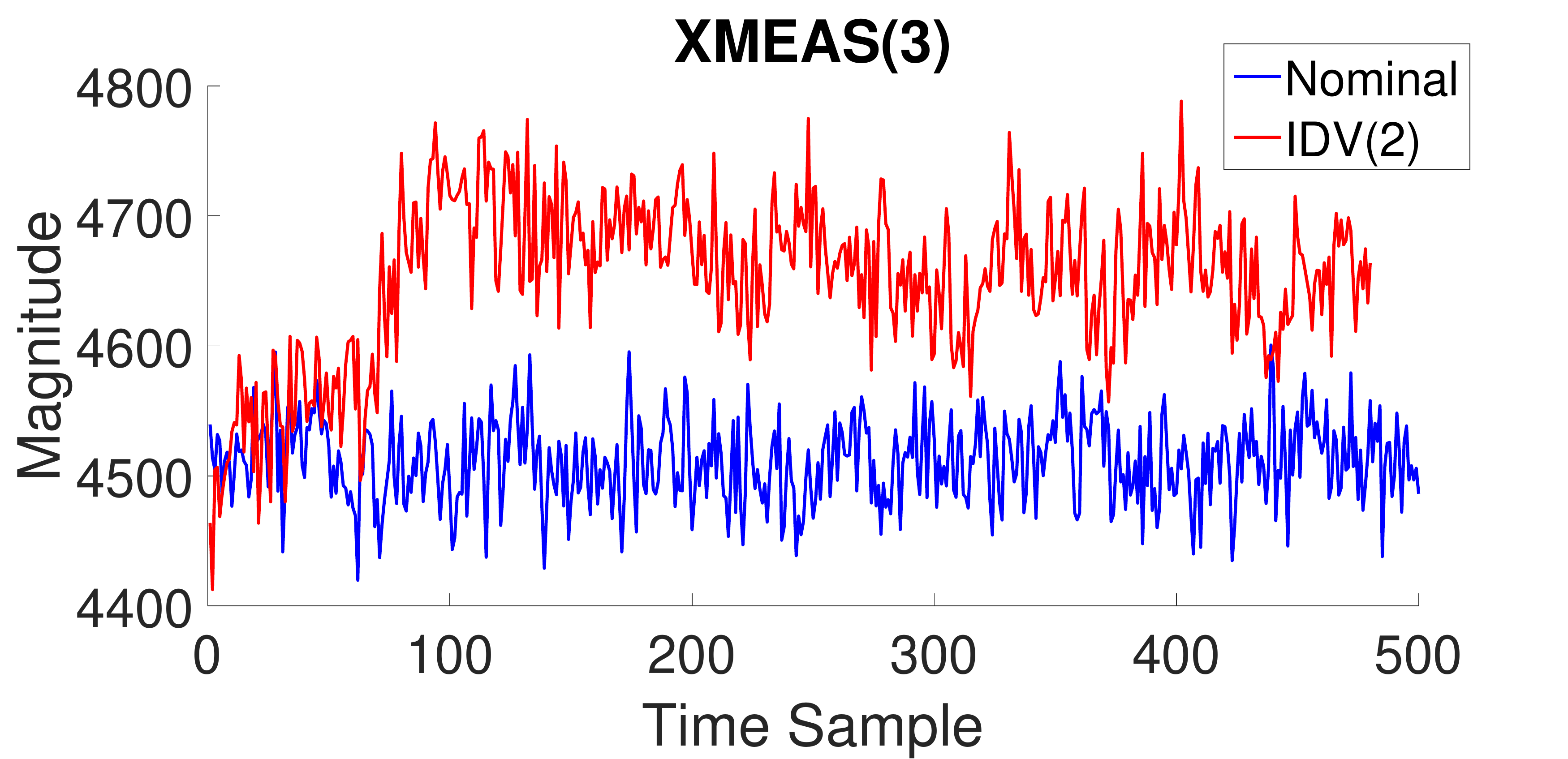}
    \includegraphics[width = 0.32\textwidth]{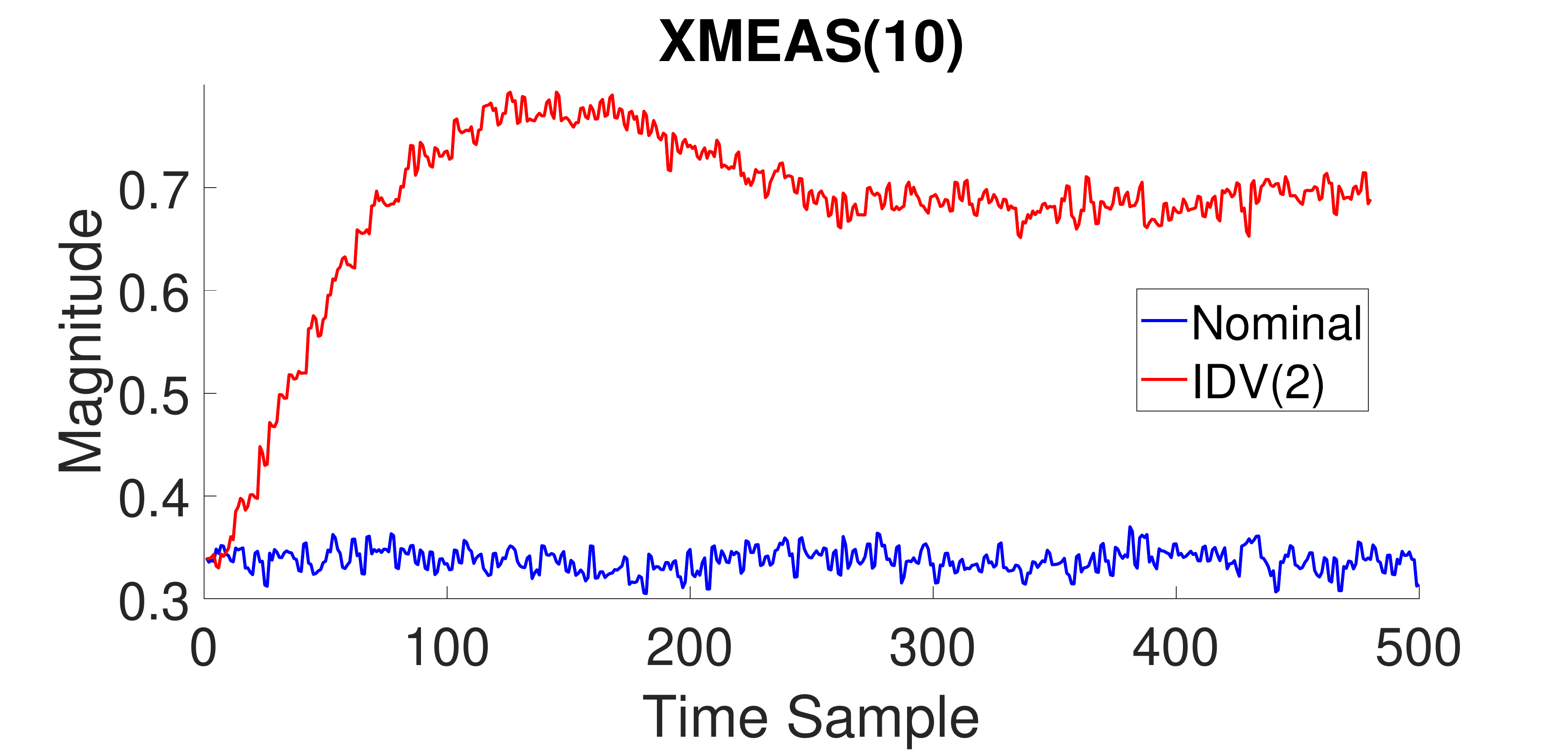}
    \includegraphics[width = 0.32\textwidth]{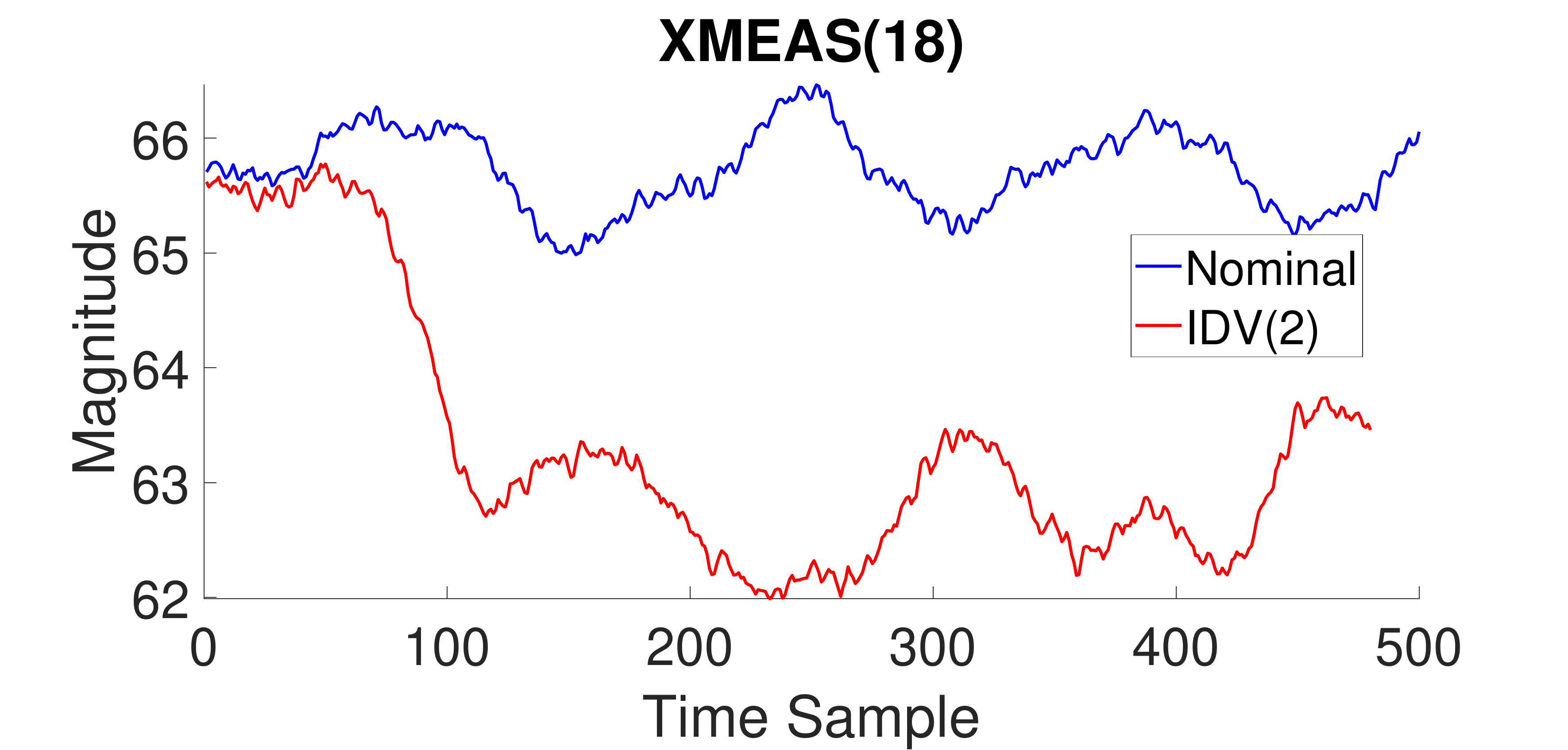}
    \includegraphics[width = 0.32\textwidth]{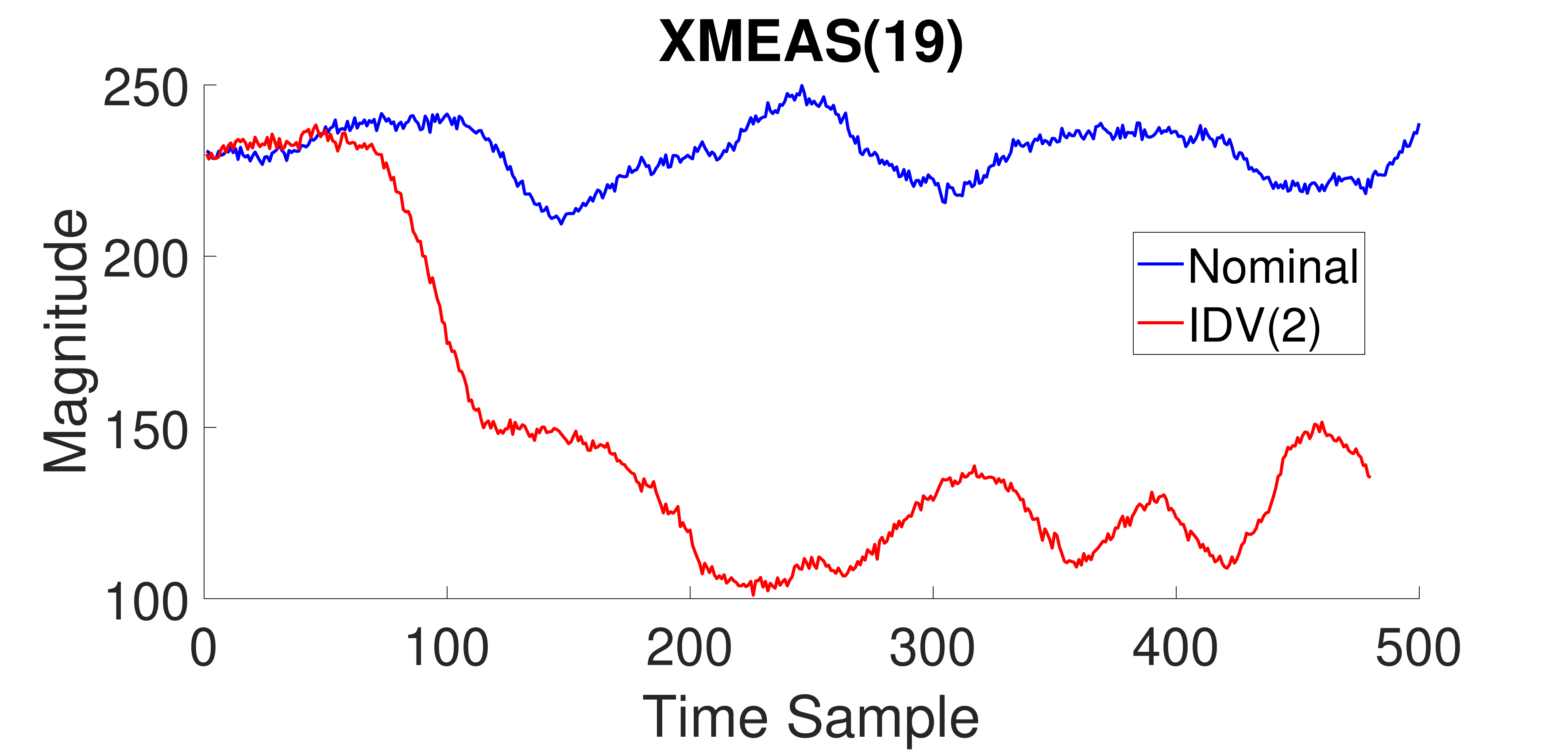}
    \includegraphics[width = 0.32\textwidth]{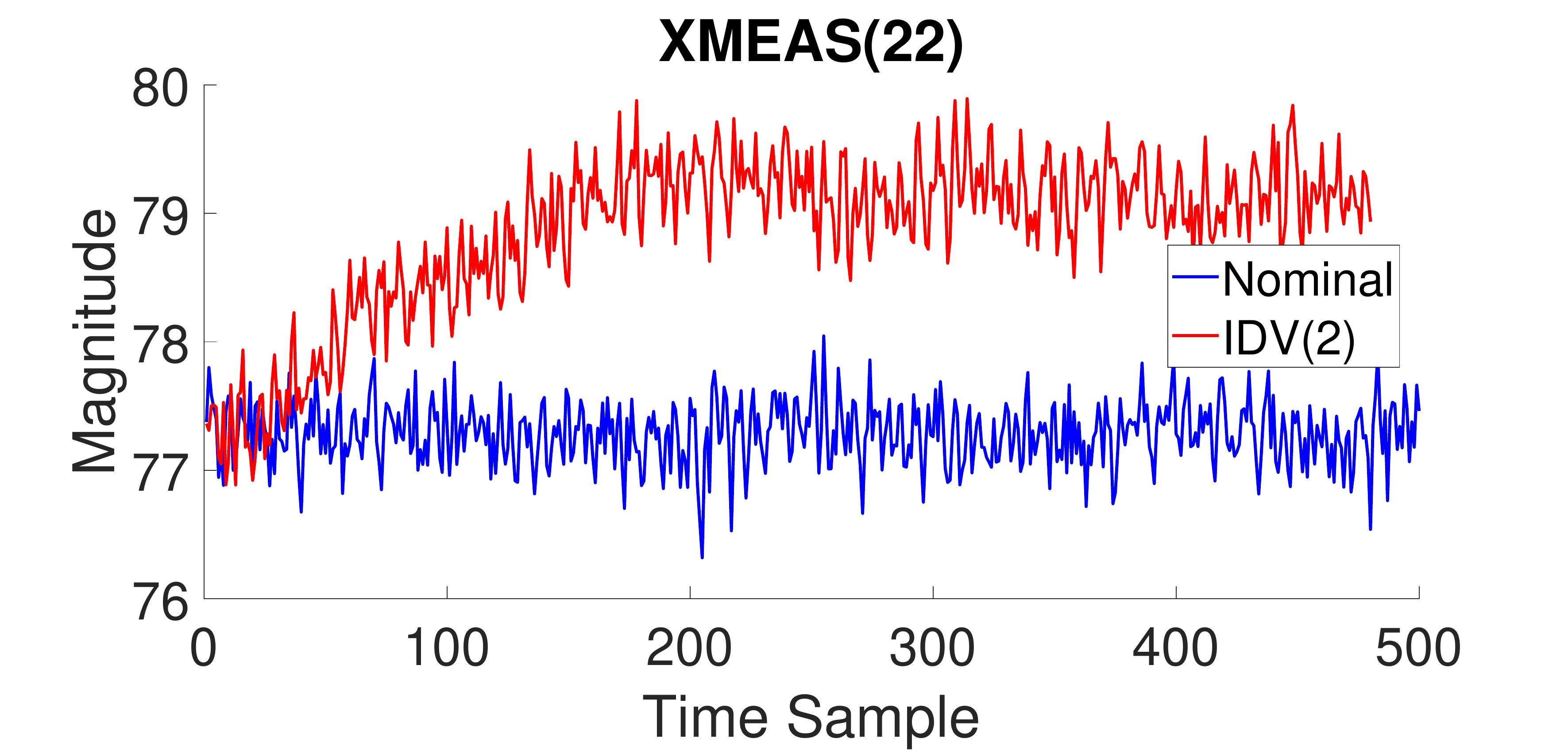}
    \includegraphics[width = 0.32\textwidth]{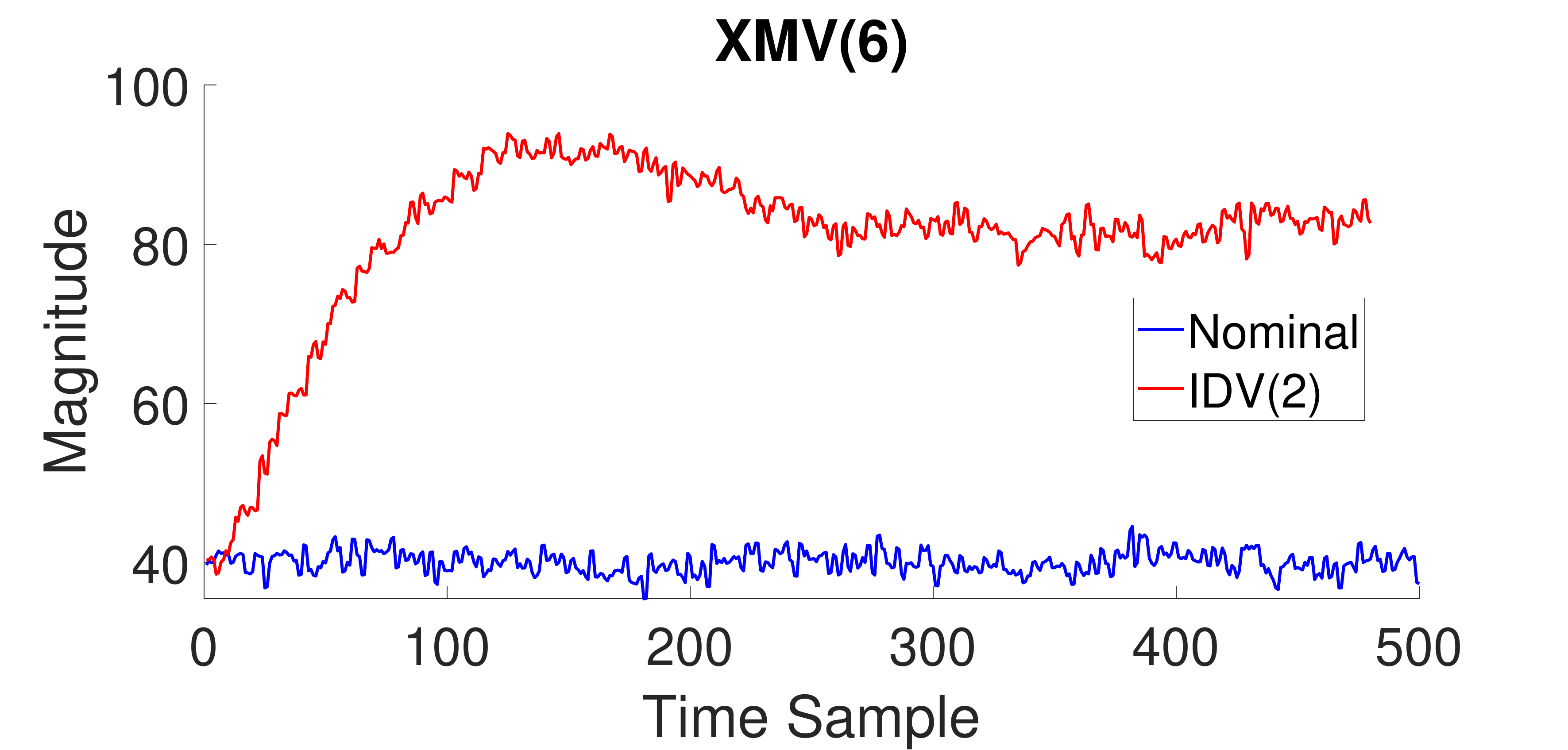}
    \includegraphics[width = 0.32\textwidth]{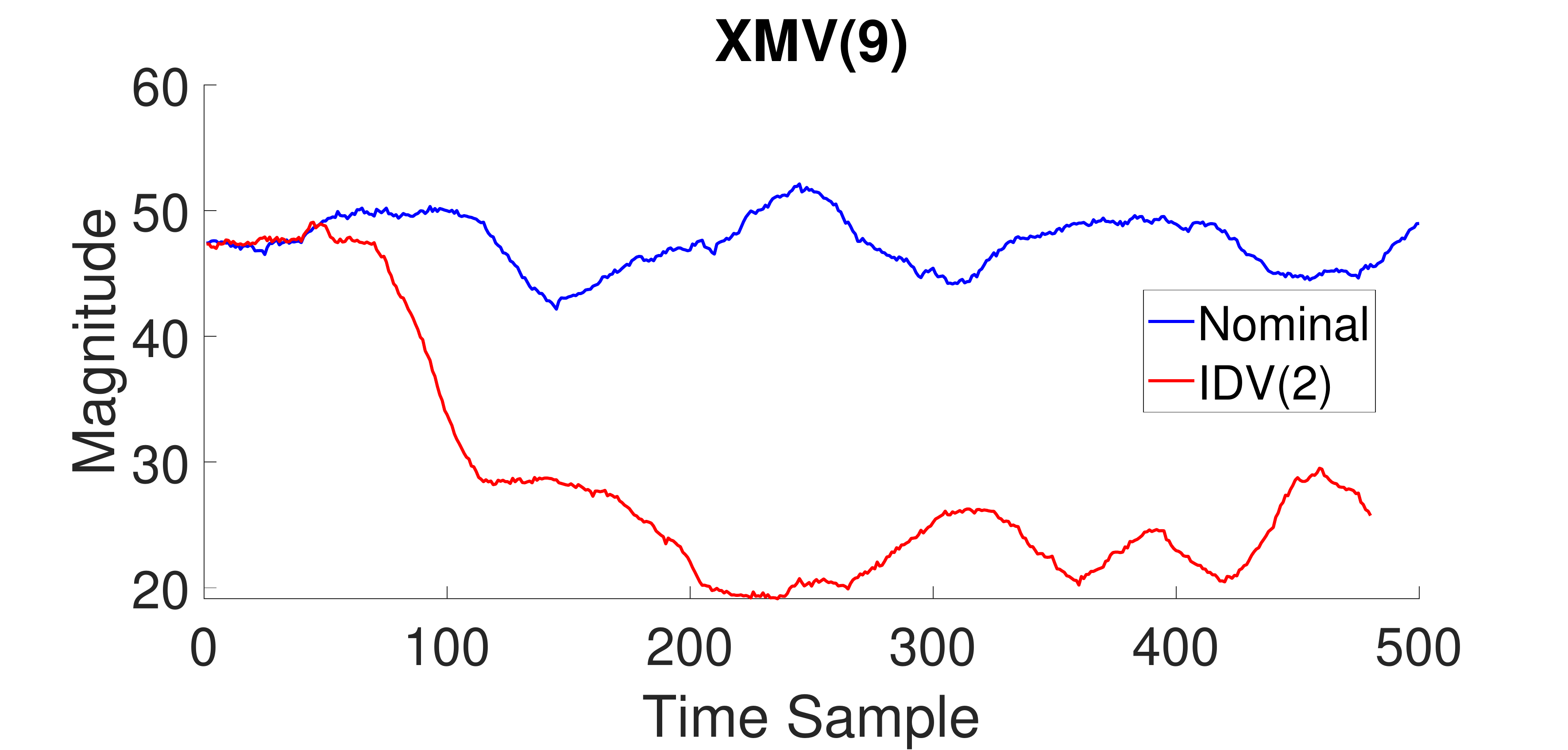}
    \includegraphics[width = 0.32\textwidth]{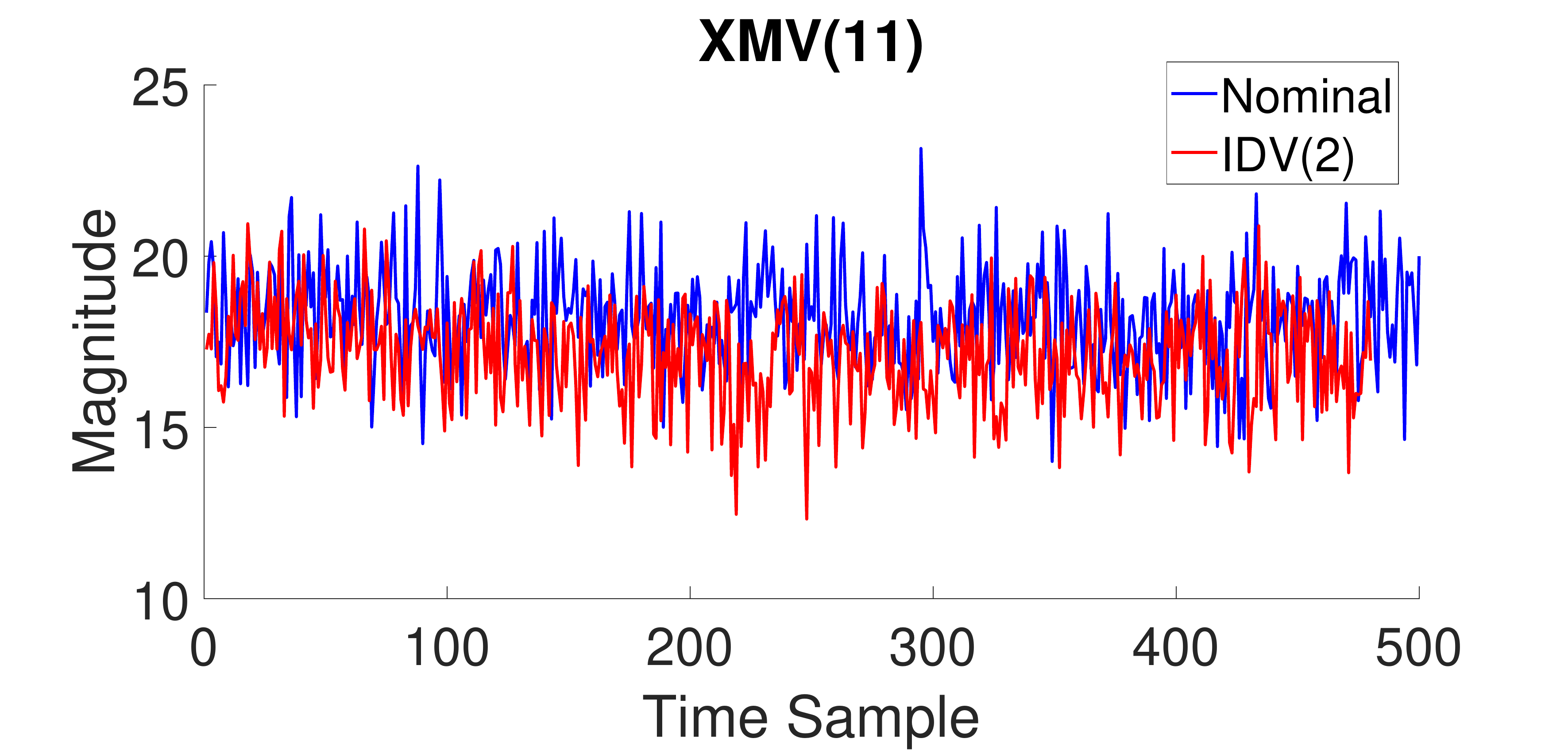}
    \caption{Relevant variables contributing to IDV(2) with nominal and abnormal profiles}
    \label{deviation_nominal_2}
\end{figure}

\begin{figure}
    \centering
    \includegraphics[width = \textwidth]{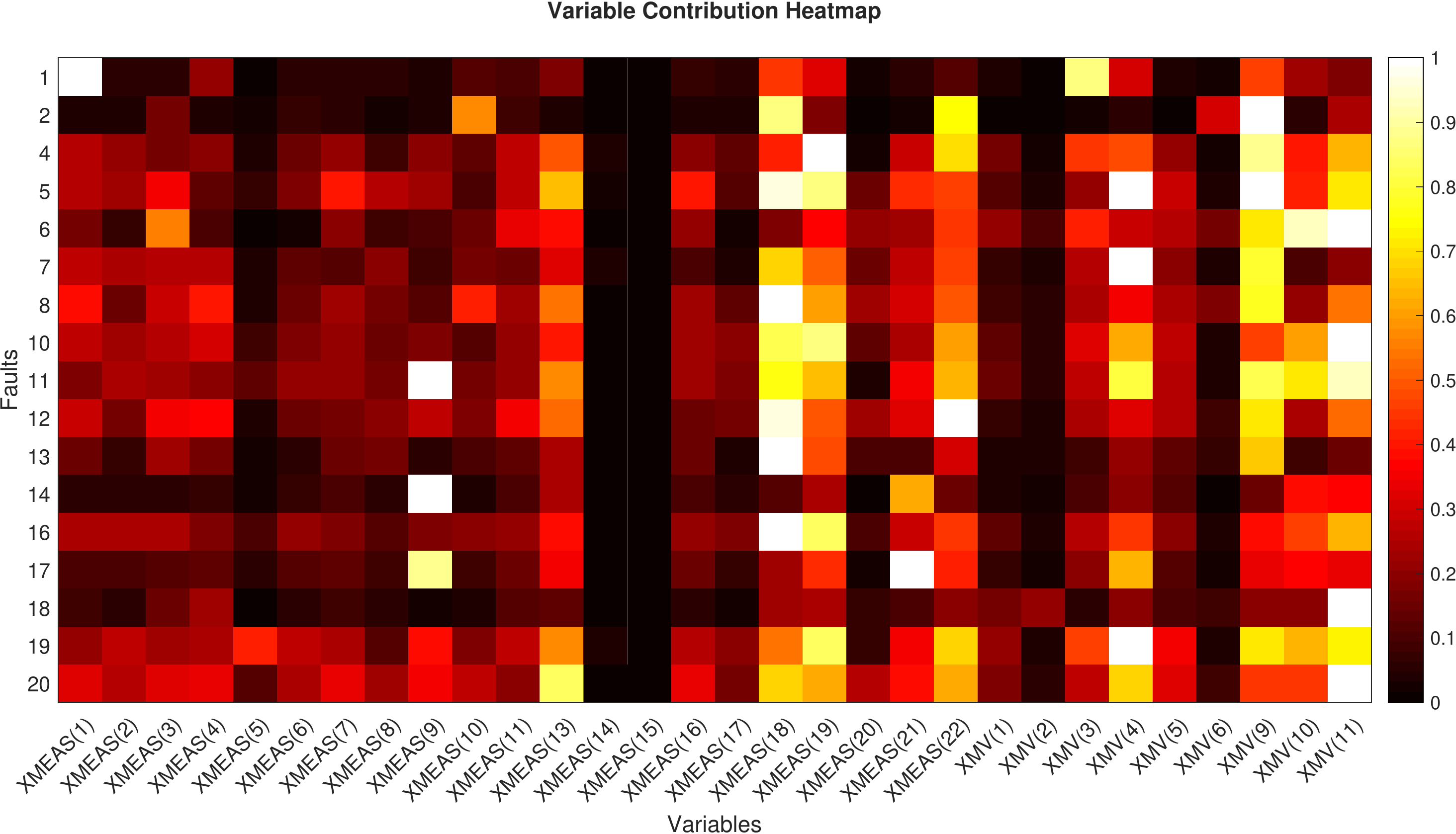}
    \caption{Variable Contribution Heatmap corresponding to all faults }
    \label{heatmap_variable_contribution}
\end{figure}

\begin{figure}[]
    \centering
    \includegraphics[width =1.1\textwidth, height = 19em]{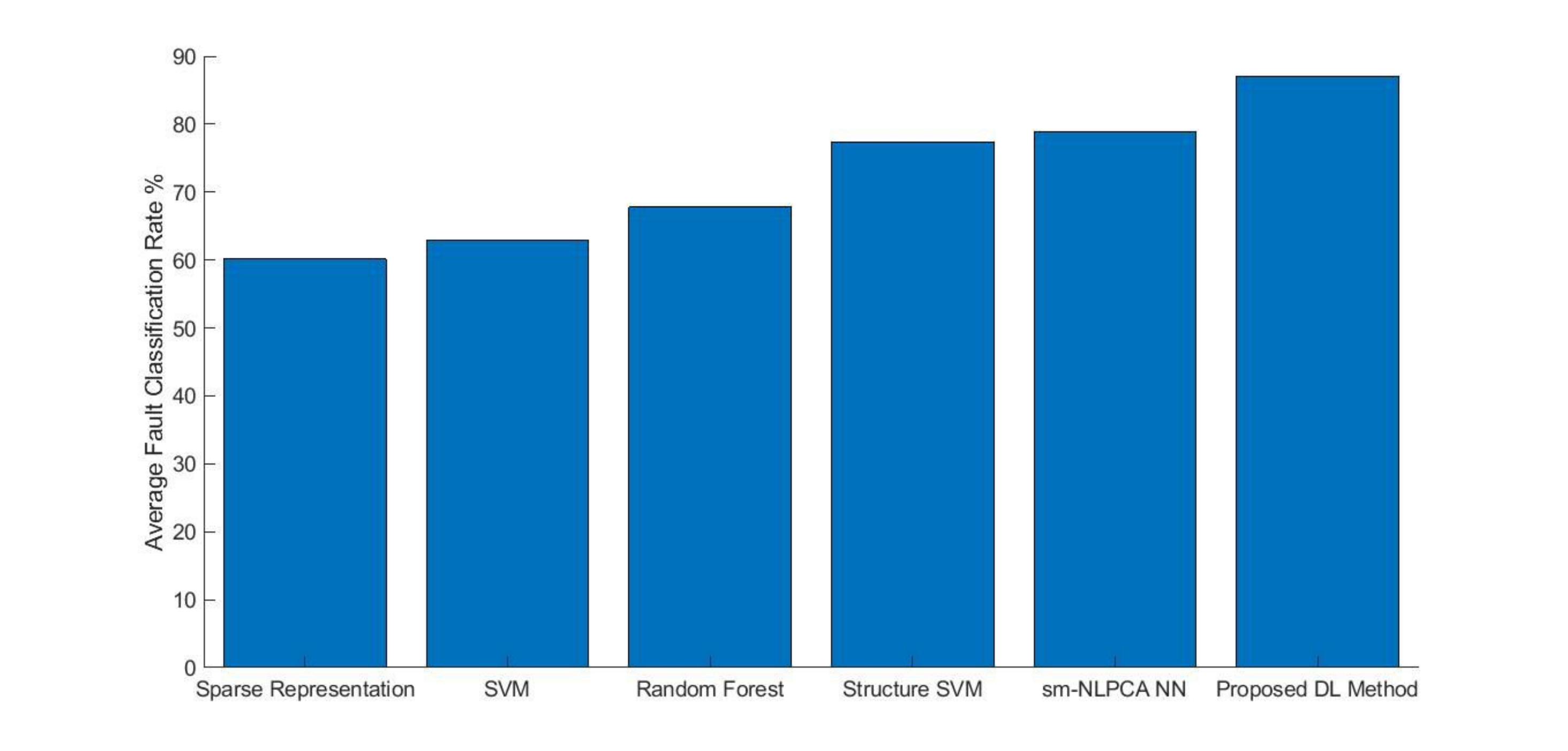}
    \caption{Comparison of Fault Classification rate with different methods}
    \label{comparison}
\end{figure}

\begin{table}[]
\caption{Comparison of Fault Detection Rate with different methods with non-incipient faults only}
\label{Fault detection comparison}
\resizebox{\textwidth}{!}{%
\begin{tabular}{@{}cccccccccccc@{}}
\toprule
\textbf{Fault} & \multicolumn{2}{c}{\textbf{\begin{tabular}[c]{@{}c@{}}PCA\\ (15 comp.)\end{tabular}}} & \textbf{\begin{tabular}[c]{@{}c@{}}DPCA\\ (22 comp.)\end{tabular}} & \multicolumn{2}{c}{\textbf{\begin{tabular}[c]{@{}c@{}}ICA\\ (9 comp.)\end{tabular}}} & \textbf{\begin{tabular}[c]{@{}c@{}}DL\\ (2017)\end{tabular}} &
\textbf{\begin{tabular}[c]{@{}c@{}}DL\\ (2017)\end{tabular}} &
\textbf{\begin{tabular}[c]{@{}c@{}}DL\\ (2018)\end{tabular}} &
\textbf{\begin{tabular}[c]{@{}c@{}}DL\\ (2018)\end{tabular}}&
\textbf{\begin{tabular}[c]{@{}c@{}}DL\\ (2019)\end{tabular}} & \textbf{\begin{tabular}[c]{@{}c@{}}Proposed \\DL \\ \end{tabular}} \\ \midrule
 & \textbf{$T^2$} & \textbf{SPE} & \textbf{$T^2$} & \textbf{$I^2$} & \textbf{AO} & {SAE-NN} & {DSN} & {GAN} & OCSVM & {CNN} &  DDSAE\\  \hline
1 & 99.2\% & 99.8\% & 99\% & 100\% & 100\% & 77.6\% & 90.8\% & 99.62\% & 99.5\%& 91.39\% & 99.9\% \\
2 & 98\% & 98.6\% & 98\% & 98\% & 98\% & 85\% & 89.6\% & 98.5\% & 98.5\% & 87.96\% & 98.2\% \\
4 & 4.4\% & 96.2\% & 26\% & 61\% & 84\% & 56.6\% & 47.6\% & 56.25\% & 50.37\% & 99.73\% & 99.7\% \\
5 & 22.5\% & 25.4\% & 36\% & 100\% & 100\% & 76\% & 31.6\% & 32.37\% & 30.5\% & 90.35\% & 100\% \\
6 & 98.9\% & 100\% & 100\% & 100\% & 100\% & 82.8\% & 91.6\% & 100\% & 100\%& 91.5\% & 99.7\% \\
7 & 91.5\% & 100\% & 100\% & 99\% & 100\% & 80.6\% & 91\% & 99.99\% & 99.62\%& 91.55\% & 99.9\% \\
8 & 96.6\% & 97.6\% & 98\% & 97\% & 97\% & 83\% & 90.2\% & 97.87\% & 97.37\% & 82.95\% & 94\% \\
10 & 33.4\% & 34.1\% & 55\% & 78\% & 82\% & 75.3\% & 63.2\% & 50.87\%& 53.25\% & 70.05\% & 89.1\% \\
11 & 20.6\% & 64.4\% & 48\% & 52\% & 70 & 75.9\% & 54.2\% & 58\% & 54.75\%& 60.16\% & 87.7\% \\
12 & 97.1\% & 97.5\% & 99\% & 99\% & 100\% & 83.3\% & 87.8\%& 98.75\% & 98.63\%& 85.56\% & 99.4\% \\
13 & 94\% & 95.5\% & 94\% & 94\% & 95\% & 83.3\% & 85.5\% & 95\% & 94.87\%& 46.92\% & 95.6\% \\
14 & 84.2\% & 100\% & 100\% & 100\% & 100\% & 77.8\% & 89\%& 100\% & 100 \%& 88.88\% & 100\% \\
16 & 16.6\% & 24.5\% & 49\% & 71\% & 78\% & 78.3\% & 74.8\%& 34.37\%& 36.37\% & 66.84\% & 94\% \\
17 & 74.1\% & 89.2\% & 82\% & 89\% & 94\% & 78\% & 83.3\% & 91.12\% & 87.25\%& 77.11\% & 97.7\% \\
18 & 88.7\% & 89.9\% & 90\% & 90\% & 90\% & 83.3\% & 82.4\% & 90.37\% & 90.12\% & 82.74\% & 90.9\% \\
19 & 0.4\% & 12.7\% & 3\% & 69\% & 80\% & 67.7\% & 52.4\%&  11.8\% & 3.75\%& 70.87\% & 89.9\% \\
20 & 29.9\% & 45\% & 53\% & 87\% & 91\% & 77.1\% & 44.1\%& 58.37\% & 52.75\% & 72.88\% & 89.6\% \\ \hline
Average & 61.77\% & 74.72\% & 72.35\% & 87.29\% & 91.70\% & 77.7\% & 76.84\% & 74.04\% &62.78\% & 85.47\% & 96.43\% \\ \bottomrule
\end{tabular}}
\label{comparison-detection}
\end{table}

For fault classification with the static DSAE-NN models it is observed that only 235 out of 363 input variables are the most relevant features for obtaining the highest testing accuracy in identifying the type of fault. Every iteration of the proposed methodology conducted for pruning of irrelevant input features increase the classification accuracy as shown in Table \ref{architecture_fault_diagnosis}. After identifying the 33 most relevant process variables with the static DSAE-NN model, the reduced input data matrix $\{\textbf{X}^{l}_r\}$ is stacked with lagged time stamps $\{\textbf{X}^{l}_r\} \rightarrow \{\textbf{X}^{l D}_r\}$ and the network is retrained. The best test classification accuracy of $88.41\%$ is achieved by stacking ten previous time-stamp process values. The confusion matrices for the first iteration and the final iteration are shown in Figure \ref{fault_classification_confusion_matrix:initial} and \ref{fault_classification_confusion_matrix:final} respectively. It can be seen that there is a significant improvement in the average test classification due to the implementation of proposed methodology. For example: there is an increase of $38\%$ in degree of separability in IDV(8) and $20\%$ increase in IDV(13).The averaged test accuracy for fault classification problem is compared with various non-linear classification algorithms such as Sparse representation, SVM, Random Forest, Structure SVM, AE based classification (sm-NLPCA) method. As shown in Figure 14 the proposed methodology outperforms other methods by a significant margin. The averaged relative importance of each relevant input feature $\textbf{R}_c$ (estimated using Equation 14) towards the classification task $c$ (fault classification) is shown in Figure \ref{fault_classification_relevance}. An important by-product of the proposed pruning methodology is that it can explain which input variables significantly deviate from their normal trajectories while the fault is occurring or to identify the root cause of the process fault. Towards that task, averaged input relevances' values for the correctly classified samples for a specific process fault are computed using LRP. For example for Fault 1 (a step change in A/C Feed ratio)  the average relative relevance plot for IDV(1) is shown in Figure \ref{relevance for Fault 1}. Then, using this plot which are the variables that will significantly deviate from their trajectories during normal operation following the occurrence of the fault.  Figure \ref{deviation_nominal_1} shows the evolution of the IDV(1) relevant input variables as a function of time. This figure corroborates that all the identified variables according to the average relevance analysis do significantly deviate from their nominal operation values during the occurrence of fault IDV(1). Similarly, an averaged relative relevance plot for IDV(2) is shown in Figure \ref{relevance for Fault 2} and the input variables identified as significant to detect this fault are then plotted in Figure \ref{deviation_nominal_2} as functions of time corroborating that the variables identified as significant in the plot \ref{relevance for Fault 2} are significantly deviating following the fault from their trajectories during normal operation.  Similar diagnostics can be run for all the other faults for the TEP problem.\\

Finally, for easier visualization, using these average relative relevances' values for each fault a variable contribution heatmap is generated and shown in Figure \ref{heatmap_variable_contribution} can be computed using Equation 14. In this heatmap the significance of different variables are color coded to identify the possible root-cause of different variables for each particular fault. For example for fault 1, variables 1 and 3 are shown as significant in the heatmap and it can be verified from Figure 9 that these same variables deviate the most during faulty operation from their trajectories during normal operation.

\section{Conclusion}
In this paper, an explainability based fault detection and classification methodology is proposed using both a deep supervised autoencoder (DSAE) and  dynamic supervised autoencoder (DDSAE) for the extraction of features. A Layerwise Relevance Propagation (LRP) algorithm is used as the main tool for explaining the classification predictions for the deep neural networks. The explainability measure serves two major objectives: i) Pruning of irrelevant input variables and further improvement in the test classification accuracy and ii) Identification of possible root cause of different faults occurring in the process.  The fault detection and classification performance of the proposed DSAE/xDSAE and DDSAE/xDSSAE DNN models together is tested on the TE benchmark process. The proposed methodology outperforms both multivariate linear methods and other DL based methods reported in the literature on the same standard data. \\

Although this study make use of the powerful feature extraction capability of deep learning neural network models and XAI (eXplainable AI) their use in industrial processes must face practical challenges such as availability of data for training and a longer development time for off-line model calibration because of the sequence of pruning and re-training steps.

\section{Acknowledgement}
This work is the result of the research project supported by MITACS grant IT10393 through MITACS-Accelerate Program.








\bibliographystyle{cas-model2-names}

\bibliography{cas-refs}



\end{document}